\documentclass[aip,pof,amsmath,amssymb,reprint]{revtex4-2}
\usepackage[utf8]{inputenc}
\usepackage{graphicx}
\usepackage{dcolumn}
\usepackage{bm}
\usepackage{xcolor}

\usepackage{float}
\restylefloat{figure}
\restylefloat{table}
\makeatletter
\renewcommand\floatc@plain[2]{\@makecaption{#1}{#2}}
\makeatother
\usepackage[T1]{fontenc}
\usepackage{etoolbox}
\usepackage{xurl}
\usepackage[hidelinks]{hyperref}
\makeatletter
\def\@email#1#2{%
 \endgroup
 \patchcmd{\titleblock@produce}
  {\frontmatter@RRAPformat}
  {\frontmatter@RRAPformat{\produce@RRAP{Corresponding author: \href{mailto:#2}{#2}}}\frontmatter@RRAPformat}
  {}{}
}
\makeatother


\begin{document}

\preprint{}

\title[]{Beyond Affine Reconstruction: Multi-Resolution Transformer DeepONets for the Radiative Transfer Equation}

\author{Tanguy Azéma}
\affiliation{CEREA, ENPC, EDF R\&D, Institut Polytechnique de Paris, \^{I}le-de-France, France}
\author{Marc Bocquet}
\affiliation{CEREA, ENPC, EDF R\&D, Institut Polytechnique de Paris, \^{I}le-de-France, France}
\author{Fatiha Nmira}
\affiliation{EDF R\&D, Chatou, France}
\author{Xiaoyi Lu}
\affiliation{Research Division, FM, 1165 Boston-Providence Turnpike,
Norwood, MA 02062, USA}
\author{Jean-Louis Consalvi}
\affiliation{Aix-Marseille Université, CNRS, IUSTI UMR 7343, Marseille, France}
\author{Sibo Cheng}
\email{sibo.cheng@enpc.fr}
\affiliation{CEREA, ENPC, EDF R\&D, Institut Polytechnique de Paris, \^{I}le-de-France, France}

\date{\today}

\begin{abstract}
Radiative heat transfer plays a dominant role in fire simulations, where solving the radiative transfer equation (RTE) is a computational bottleneck. Neural operators can accelerate these computations by learning solution operators from offline simulation data. However, conventional DeepONet surrogates rely on fixed-grid inputs and affine reconstruction, limiting their flexibility across resolutions and their approximation accuracy at a fixed latent dimension. In this paper, we develop a transformer-based DeepONet for multi-resolution RTE surrogate modelling. The proposed approach combines a Perceiver encoder, which maps temperature and absorption coefficient fields at different spatial resolutions into a fixed-size latent representation, with a cross-attention decoder that enables non-affine reconstruction of the radiative intensity. This architecture allows a single model to process inputs at varying resolutions without resampling them onto a common grid. We formulate a computable empirical version of a previously derived theoretical lower bound for affine reconstruction, enabling direct comparison with model errors on the same dataset. Experiments on a two-dimensional pool-fire dataset demonstrate that the proposed approach outperforms autoencoder-based DeepONets in predicting both radiative intensity and incident radiation, with fewer trainable parameters. Power spectral density analysis shows improved preservation of small-scale structures. Both affine baselines remain above the corresponding lower bound at all evaluated ranks, whereas the proposed model achieves a training error below it at one matched latent dimension, demonstrating the potential of non-affine reconstruction to overcome this representational limitation. Experiments on a dataset combining three spatial resolutions confirm the accuracy advantage of the approach at each input resolution.
\end{abstract}


\maketitle

\section{\label{sec:intro}Introduction}

Radiation is a major mode of heat transfer in many fire scenarios, and accurately modelling radiative heat exchange is essential for predicting fire growth. In coupled fire simulations, the radiative source term in the energy equation is evaluated by solving the radiative transfer equation (RTE), which governs the spatial and directional distribution of radiative intensity. Repeatedly solving the RTE using conventional numerical methods, such as the finite angle or discrete ordinates method~\cite{modest2021radiative}, can be computationally expensive because both physical space and angular directions must be discretised. Coarsening these discretisations reduces the computational cost but can compromise accuracy~\cite{chai1993ray}. Conversely, converged spatial and angular discretisations increase the computational burden substantially, making such simulations unaffordable in most applications~\cite{kim2023evaluation}. This trade-off between accuracy and computational efficiency motivates the development of fast surrogate models capable of preserving the spatial and angular structure of radiative fields.

Similar challenges arise in many partial differential equation (PDE) problems, motivating the growing use of deep learning to construct surrogate models that provide rapid predictions after offline training. Conventional approaches include supervised learning on generated input--output pairs for surrogate modelling~\cite{hesthaven2018nonintrusive} and direct training with PDE-based losses, as in the physics-informed neural network (PINN) paradigm~\cite{raissi2019physics}. Both approaches may have limited generalisability beyond their training regimes, particularly to unseen parameter ranges or boundary conditions~\cite{takamoto2022pdebench}. More recent efforts have focused on learning PDE solution operators directly between function spaces, allowing a single model to represent a family of PDE solutions and giving rise to neural operators such as DeepONet and the Fourier neural operator~\cite{lu2021learning,li2021fourier}. This formulation is particularly attractive for coupled flow simulations, where the same learned operator can be repeatedly evaluated as the input fields evolve, replacing repeated numerical solutions of the RTE with forward passes through a trained network.

Applications of neural operators to the RTE have emerged only recently. Lu and Wang~\cite{lu2024surrogate} introduced a physics-informed MIONet for nongray gas radiative heat transfer, while subsequent work applied DeepONet to accelerate RTE calculations in computational fluid dynamics (CFD) fire simulations~\cite{lu2026machine}. Fourier-based neural operators have also been investigated for two- and three-dimensional fire configurations and prediction across spatial discretisations~\cite{carne2026fourier}. Despite this progress, how the reconstruction mechanism affects the accuracy of RTE surrogates remains poorly understood. In particular, current approaches~\cite{lu2024surrogate,lu2026machine} are based on standard DeepONets, which employ affine reconstruction and are therefore subject to a rank-dependent lower bound on the approximation error~\cite{lanthaler2022error}. Furthermore, operational RTE simulations commonly involve multi-resolution or locally refined meshes~\cite{qu2011micrometeorological,nmira2020influence}, whereas existing approaches assume fixed-grid input snapshots, limiting their applicability to industrial workflows.

Both limitations are structural to the DeepONet in its original formulation~\cite{lu2021learning}. Its architecture consists of two components: a branch network that encodes the input function and a trunk network that maps output coordinates to basis functions. The prediction is then obtained as an affine combination of these basis functions, with coefficients provided by the branch network. The first limitation arises from the branch network, which takes input values at a fixed set of sensor locations, requiring fields defined on a different mesh to be resampled before evaluation. This requirement limits the flexibility of surrogates intended to replace numerical solvers in simulations involving varying mesh resolutions.
The second limitation arises from the reconstruction, which is affine in the branch coefficients and therefore subject to the approximation error lower bound established in Theorem 3.4 of Lanthaler \emph{et al.}~\cite{lanthaler2022error}. We define the rank of a DeepONet as the number of coefficients produced by the branch network, equal to the number of trunk functions used in the reconstruction. This quantity can also be interpreted as the dimension of the latent representation of the input function passed to the reconstructor. At a fixed reconstruction rank, the approximation error of any model in this class is subject to a theoretical lower bound that cannot be overcome by modifying the input encoding, increasing the network width, or extending training. This bound can be evaluated separately for the empirical measures of the training and test datasets, providing a corresponding lower limit on the error for each. For a given solution operator and evaluation measure, the bound depends only on the reconstruction rank and can decrease only as the rank increases.
Previous studies have derived analytical estimates of this bound for benchmark problems~\cite{lanthaler2022error,lanthaler2023nonlinear} and used empirical principal component analysis (PCA) spectra to motivate nonlinear decoders~\cite{seidman2022nomad}. However, to the best of our knowledge, this bound has not been explicitly evaluated under the empirical measure of a complex physical simulation dataset and directly compared with model errors computed on the same dataset.

Overcoming this representational limitation requires moving beyond affine reconstruction. Attention mechanisms provide a flexible framework for this purpose, enabling decoders in which predictions depend nonlinearly on the encoded input. They can also map inputs of varying spatial resolutions to a fixed-size latent representation, enabling multi-fidelity training and inference without interpolation.
Combining these two capabilities offers a way to address both the reconstruction bottleneck and the fixed-grid input requirement of conventional DeepONets.
Although attention mechanisms have advanced neural operators such as OFormer and Continuous Vision Transformer (CViT)~\cite{li2023oformer,wang2025cvit}, their application to radiative transfer remains limited. For instance, while DeepRTE employs attention for scattering media, it still relies on an affine reconstruction~\cite{zhu2026deeprte}. Furthermore, although Fourier-based operators have successfully processed multi-resolution data across various physical systems~\cite{carne2026fourier}, the surrogate models currently coupled to fire solvers are strictly restricted to affine DeepONets~\cite{lu2026machine}. To the best of our knowledge, this work presents the first attention-based neural operator for fire radiative transfer. More crucially, none of these existing RTE affine surrogates have been evaluated against the theoretical lower error bound that mathematically governs their class. Our approach bridges this gap by introducing the first attention-based operator for fire radiation, uniquely leveraging a non-affine reconstructor to overcome this fundamental representational bottleneck while natively handling multi-resolution inputs.
The main contributions of this paper could be summarised as:

\begin{itemize}
\item  We develop a transformer-based neural operator for the RTE that processes regular-grid inputs at varying spatial resolutions without the explicit resampling required by the DeepONet surrogates currently used for fire radiation.

\item  We systematically compare the proposed model with affine DeepONet baselines for fire-radiation modelling and show that it consistently outperforms both PCA- and autoencoder-DeepONet on single- and multi-resolution datasets.

\item  We evaluate, for the first time to our knowledge, the theoretical lower bound derived by Lanthaler \emph{et al.}~\cite{lanthaler2022error} on the empirical measure of a complex, real-world PDE dataset. Both affine DeepONet baselines remain above this bound, whereas the transformer-based model achieves an error below it at one rank, showing that nonaffine reconstruction can overcome this representational bottleneck.
\end{itemize}

The rest of this paper is organised as follows. Sec.~\ref{sec:related} introduces the radiative transfer equation, the reference dataset, and the PCA- and autoencoder (AE)-DeepONet baselines. Sec.~\ref{sec:method} details the proposed transformer-based DeepONet and formulates the empirical lower bound for affine reconstructions. Sec.~\ref{sec:results} presents the numerical experiments, evaluating predictive accuracy, comparing the baselines against the theoretical error bound, and assessing generalisation across spatial resolutions. Finally, Sec.~\ref{sec:conclusion} concludes the paper.

\begin{figure}[htbp]
  \makebox[\textwidth][c]{\includegraphics[width=1.0\textwidth]{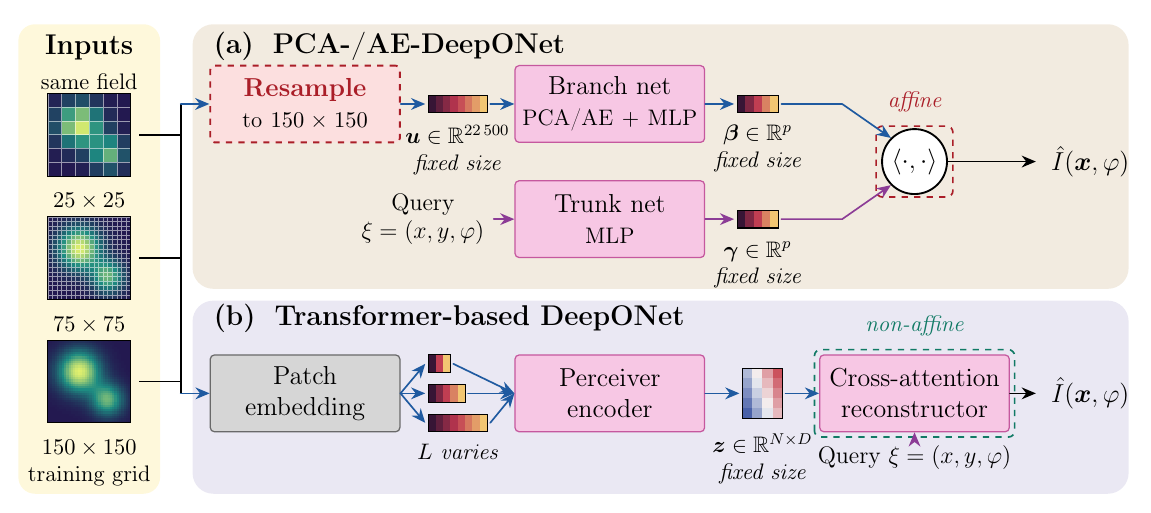}}
  \caption{Comparison of encoding and reconstruction architectures. (a): In PCA- and AE-DeepONet baselines, the input sensor grid and the branch/trunk output widths ($p$) are strictly fixed. Consequently, multi-resolution inputs require explicit resampling, and the final inner product yields an affine reconstruction constrained by a theoretical error bound. (b): The proposed transformer-based DeepONet dynamically adapts the number of patch tokens ($L$) to the input resolution while maintaining a fixed-size latent state ($z \in \mathbb{R}^{N \times D}$), natively handling varying resolutions without resampling. Furthermore, its cross-attention reconstructor evaluates the query as a non-affine function of the latent state, overcoming the affine representational bottleneck. Both models evaluate the intensity pointwise given the same query $\xi=(x,y,\varphi)$.}
  \label{fig:resolution}
\end{figure}

\section{\label{sec:related}Problem setting and related works}

\subsection{\label{subsec:setup}Problem setting, solution operator and reference data}
We consider a gray, absorbing and emitting medium, and neglect scattering. In two space dimensions, the radiative intensity $I(\bm{x},\varphi)$ then solves the radiative transfer equation (RTE)
\begin{equation}
  \label{eq:rte}
  \bm{s}(\varphi)\cdot\nabla I(\bm{x},\varphi)
  =\kappa(\bm{x})\big(I_b(\bm{x})-I(\bm{x},\varphi)\big),
  \qquad \bm{x}\in\mathcal{D}\subset\mathbb{R}^{2},\ \varphi\in[0,2\pi).
\end{equation}

The variables are the following. The position $\bm{x}=(x,y)$ runs over the two-dimensional spatial domain $\mathcal{D}\subset\mathbb{R}^{2}$, and the direction of propagation is carried by the single angle $\varphi$, measured clockwise from the $y$ axis, the associated unit vector being $\bm{s}(\varphi)=(\sin\varphi,\cos\varphi)\in\mathbb{S}^{1}$. The problem thus involves a single angular degree of freedom and two spatial coordinates, yielding a total of three degrees of freedom. For all neural operators evaluated in this study, the query is defined as the scalar triplet
\begin{equation}
  \label{eq:query}
  \xi=(x,y,\varphi)\ \in\ \mathcal{Q}=\mathcal{D}\times[0,2\pi),
\end{equation}
parameterising the direction using the angle² $\varphi$ itself rather than the unit vector $\bm{s}$. This formulation departs from that of Lu \emph{et al.}~\cite{lu2026machine,lu2024surrogate}, whose trunk network processes four scalars: the two spatial coordinates and the two direction cosines. We keep the query in $\mathbb{R}^{3}$: all DeepONet models considered in the present paper then read the same query, in the same format and in the same units, and the reconstructor of Sec.~\ref{subsec:fae} embeds it with random Fourier features, i.e.\ sines and cosines of the rescaled query at random Gaussian frequencies, so that the dependence on $\varphi$ is learned through these features rather than imposed through $\cos\varphi$ and $\sin\varphi$. The two input fields are the absorption coefficient $\kappa:\mathcal{D}\to\mathbb{R}_{+}$, in $\mathrm{m^{-1}}$, and the temperature $T:\mathcal{D}\to\mathbb{R}_{+}$, in Kelvin; $T$ enters Eq.~\eqref{eq:rte} through the blackbody intensity $I_b=\sigma T^{4}/\pi$, with the Stefan--Boltzmann constant $\sigma=5.67\times10^{-8}\,\mathrm{W\,m^{-2}\,K^{-4}}$. The model output is the intensity $I:\mathcal{Q}\to\mathbb{R}_{+}$, in $\mathrm{W\,m^{-2}\,sr^{-1}}$.

The quantity reported in Sec.~\ref{sec:results} alongside the intensity is the incident radiation, obtained by integrating $I$ over all directions,
\begin{equation}
  \label{eq:incident_radiation}
  G(\bm{x}) = \int_{4\pi} I(\bm{x},\bm{\Omega})\,\mathrm{d}\Omega
  = 2\int_{0}^{2\pi} I(\bm{x},\varphi)\,\mathrm{d}\varphi
  \ \approx\ \frac{4\pi}{16}\sum_{m=0}^{15} I(\bm{x},\varphi_m),
\end{equation}
in $\mathrm{W\,m^{-2}}$, where $\bm{\Omega}$ is a direction of the unit sphere and $\theta$ its polar angle from the axis normal to $\mathcal{D}$. The second member identifies $I(\bm{x},\bm{\Omega})$ with the in-plane intensity of Eq.~\eqref{eq:rte}, taken as independent of $\theta$, hence the factor $\int_{0}^{\pi}\sin\theta\,\mathrm{d}\theta = 2$.
The last member is the quadrature of the reference solver, the finite-volume discrete-ordinate method (fvDOM), which carries $16$ discrete ordinates $\varphi_m$ in the plane of $\mathcal{D}$, evenly spaced by $\Delta\varphi = \pi/8$, and gives each the solid-angle weight $4\pi/16$.

In this paper we work on the two-dimensional pool fire of Lu \emph{et al.}~\cite{lu2026machine}, the \texttt{smallPoolFire2D} case of the OpenFOAM tutorial suite, run with the fire solver FireFOAM. The reference intensities are computed by fvDOM on the domain $\mathcal{D}=(-1.5,1.5)\times(0,3)\,\mathrm{m^{2}}$, discretised by a uniform $150\times150$ mesh, and on the $16$ discrete ordinates $\varphi_m$ of Eq.~\eqref{eq:incident_radiation}; the absorption coefficient comes from the gray mean model of OpenFOAM. To ensure direct comparability with the reference study, identical solver settings, mesh configurations, and angular resolutions were adopted. It is important to note that the machine learning errors are measured in this study against this numerical reference, which is itself subject to spatial and angular discretisation errors.

The queries at which the models are trained and evaluated are exactly the points at which the solver stores its solution, the $150\times150$ cell centres crossed with the $16$ discrete ordinates $\varphi_m$, each handed to the network as a query $\xi=(x,y,\varphi)$ in the format of Eq.~\eqref{eq:query}.

Each snapshot is a triplet $(T^{(n)},\kappa^{(n)},I^{(n)})$: the two input fields carry $150\times150=22\,500$ values each, and the intensity carries $22\,500\times16=3.6\times10^{5}$ values. The $6001$ snapshots are split randomly into $4200$ training, $900$ validation and $901$ test samples. Radiation being instantaneous on the time scale of the fire, each snapshot is an independent steady problem with its own pair of fields. Here, $t$ serves solely as a sample index rather than a temporal variable of Eq.~\eqref{eq:rte}; consequently, the learning objective targets a parameter-to-solution operator rather than a time-dependent trajectory. This dataset forms the foundation for training the three surrogate models and evaluating the lower bound formulated in Sec.~\ref{subsec:floor}.

The RTE thus defines a solution operator, 
\begin{equation}
  \label{eq:operator}
  \mathcal{G}:\ C(\mathcal{D})\times C(\mathcal{D})\ \longrightarrow\
  C(\mathcal{Q}),
  \qquad (T,\kappa)\ \longmapsto\ I,
\end{equation}
and which takes a pair of scalar fields in $C(\mathcal{D})^{2}$ to a
function of $C(\mathcal{Q})$. The operator is well defined because the boundary condition is the same for all the samples in the dataset. This operator constitutes the target mapping approximated by all neural operators in this study, aligning with the formulation previously investigated by Lu et al.~\cite{lu2026machine,lu2024surrogate}.

\subsection{\label{subsec:baselines}PCA- and AE-DeepONet: two encoders behind the same affine reconstructor}

\begin{figure}[H]
  \makebox[\textwidth][c]{%
    \begin{minipage}{1.0\textwidth}
      \textbf{(a)\; PCA-DeepONet}\par\smallskip
      \includegraphics[width=\linewidth]{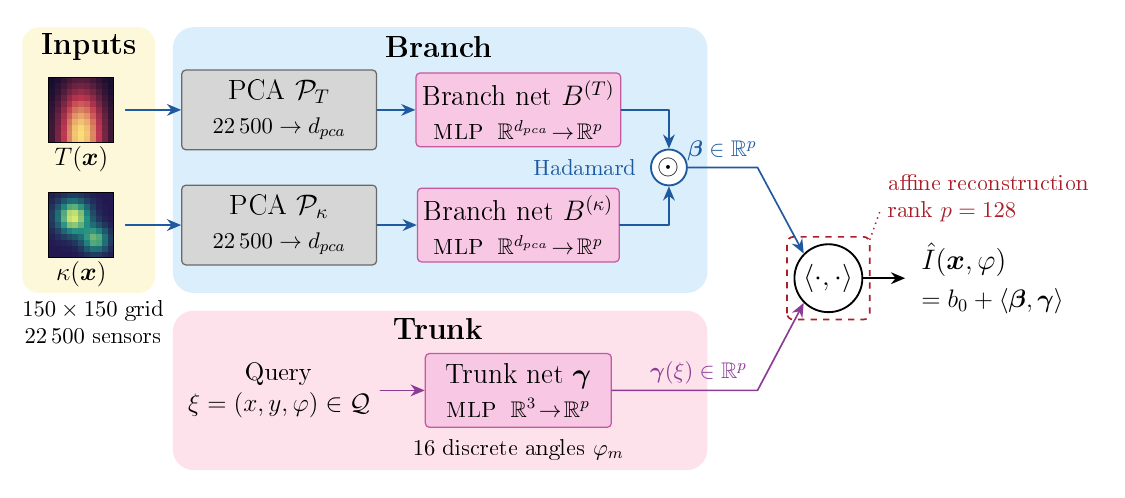}
    \end{minipage}}
  \par\medskip
  \makebox[\textwidth][c]{%
    \begin{minipage}{1.0\textwidth}
      \textbf{(b)\; AE-DeepONet}\par\smallskip
      \includegraphics[width=\linewidth]{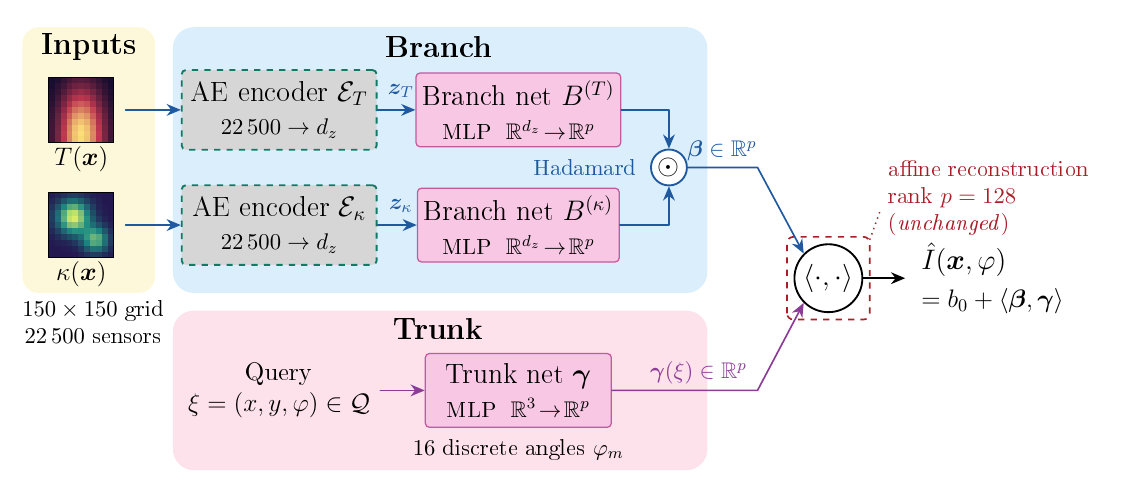}
    \end{minipage}}
  \caption{\label{fig:baselines}The two baselines. (a) PCA-DeepONet. Each input field is projected on a PCA basis learned beforehand on the training snapshots, which reduces it from $22\,500$ values to $r$ coefficients, $r$ being set by an energy criterion (Sec.~\ref{subsubsec:reconstruct}). Two branch nets and a Hadamard product turn these coefficients into a coefficient vector $\bm{\beta}\in\mathbb{R}^{p}$, paired with the trunk output by an inner product and a scalar bias. (b) AE-DeepONet. Same architecture, with the PCA replaced by the encoders of two convolutional autoencoders pretrained on $T$ and $\kappa$. Only the encoding column changes; the branch--trunk junction is left untouched. In both models the final junction is affine in $\bm{\beta}$ and of rank $p$.}
\end{figure}

Our first baseline is inspired by the PCA-DeepONet of Sorokin \emph{et al.}~\cite{sorokin2024neural}, the model Lu \emph{et al.}~\cite{lu2026machine} later coupled to a fire solver. It is a multiple-input DeepONet~\cite{jin2022mionet}: two branch nets read the temperature and absorption fields through a PCA reduction and are fused by a Hadamard product. In fact, coupling PCA to operator learning has been previously proposed by Bhattacharya \emph{et al.}~\cite{bhattacharya2021model}. The method is computed in two successive stages. The first is a reconstruction: a PCA is fitted on the training snapshots of each input field taken separately, so that $T$ and $\kappa$ receive their own basis and each is reduced from the $22\,500$ grid values to a latent vector of a chosen dimension~\cite{lu2026machine}. The PCA projection is kept fixed throughout the second stage, in which a standard DeepONet~\cite{lu2021deeponet} is trained on the resulting codes. The outputs of two branch networks are combined by a Hadamard product to form a coefficient vector, while a trunk network takes the query $\xi=(x,y,\varphi)\in\mathbb{R}^{3}$ as input. The inner product of the coefficient vector and the trunk output yields the pointwise prediction at any spatial position and angular direction as shown in Fig.~\ref{fig:baselines}(a). Finally, with $\mathcal{P}_T$ and $\mathcal{P}_\kappa$ the two frozen PCA encoders, $B^{(T)}$ and $B^{(\kappa)}$ the two branch nets and $\xi$ the query \eqref{eq:query},
\begin{subequations}
\begin{align}
  \label{eq:pcabranchT}
  \bm{\beta}_T      &= B^{(T)}\big(\mathcal{P}_T[T]\big)\in\mathbb{R}^{p},\\
  \label{eq:pcabranchK}
  \bm{\beta}_\kappa &= B^{(\kappa)}\big(\mathcal{P}_\kappa[\kappa]\big)\in\mathbb{R}^{p},\\
  \label{eq:pcatrunk}
  \bm{\gamma}(\xi)  &= \big(\gamma_1(\xi),\dots,\gamma_p(\xi)\big)\in\mathbb{R}^{p}, \\
  \bm{\beta} &=\bm{\beta}_T\odot\bm{\beta}_\kappa,
\end{align}
with $\odot$ the Hadamard product, we have
\begin{equation}
  \label{eq:pcaoutput}
  \hat{I}(\bm{x},\varphi)=b_0+\big\langle\bm{\beta},\bm{\gamma}(\xi)\big\rangle
                         =b_0+\sum_{l=1}^{p}\beta_l\,\gamma_l(\xi),
\end{equation}
where $b_0$ is a scalar bias.
\end{subequations}

Our second baseline investigates whether nonlinear dimensionality reduction can improve on PCA. PCA minimises the mean-squared reconstruction error among linear subspace approximations of a given dimension. A convolutional autoencoder can instead learn a nonlinear representation, potentially preserving more field information at the same latent dimension. We pretrain separate convolutional autoencoders for $T$ and $\kappa$ using the same snapshots as for PCA, then retain only their encoders to replace the corresponding PCA projections as illustrated in Fig.~\ref{fig:baselines}(b). Unlike PCA, these encoders could be held fixed during the second stage or fine-tuned with the branch and trunk nets. Only the encoders are replaced while the remaining DeepONet architecture, including the branch--trunk combination, is unchanged. Appendix~\ref{app:hyper} details the layers of both baselines and their two-stage training.

Both baselines share two limitations. Their encoders require a fixed input grid, so fields at a different resolution must first be resampled onto that grid. Moreover, reconstruction is affine in the branch coefficients (Eq.~\eqref{eq:pcaoutput}), restricting predictions to an affine space of dimension at most $p$. This affine reconstruction structure imposes a theoretical lower bound on the reconstruction error for a given training or validation dataset, which we compute in Sec.~\ref{subsec:floor}. The bound depends only on the reconstruction space and is independent of the encoder.

\section{\label{sec:method}Methodology}

\subsection{\label{subsec:fae}Transformer-based DeepONet: a resolution-agnostic encoder and a nonaffine reconstructor}

To overcome the aforementioned limitations of conventional DeepONets, we propose a novel transformer-based architecture designed to natively handle arbitrary spatial resolutions while performing non-affine reconstructions. The architecture rests on two primary components, namely a resolution-agnostic encoder and a continuous, non-affine reconstructor. Specifically, we integrate a Perceiver module~\cite{jaegle2021perceiver} as the encoder, which maps input temperature and absorption coefficient fields, sampled on arbitrary grids, into a fixed-size latent representation. This design avoids the quadratic computational complexity associated with full self-attention over high-resolution grids by employing a small set of learned latent tokens that attend to the input patch sequence. For the decoder, we adopt a continuous, CViT-inspired reconstructor~\cite{wang2025cvit} where query coordinates attend directly to the encoded latent features, thereby yielding a non-affine prediction of the radiative intensity.

\begin{figure}[htbp]
  \makebox[\textwidth][c]{\includegraphics[width=1.0\textwidth]{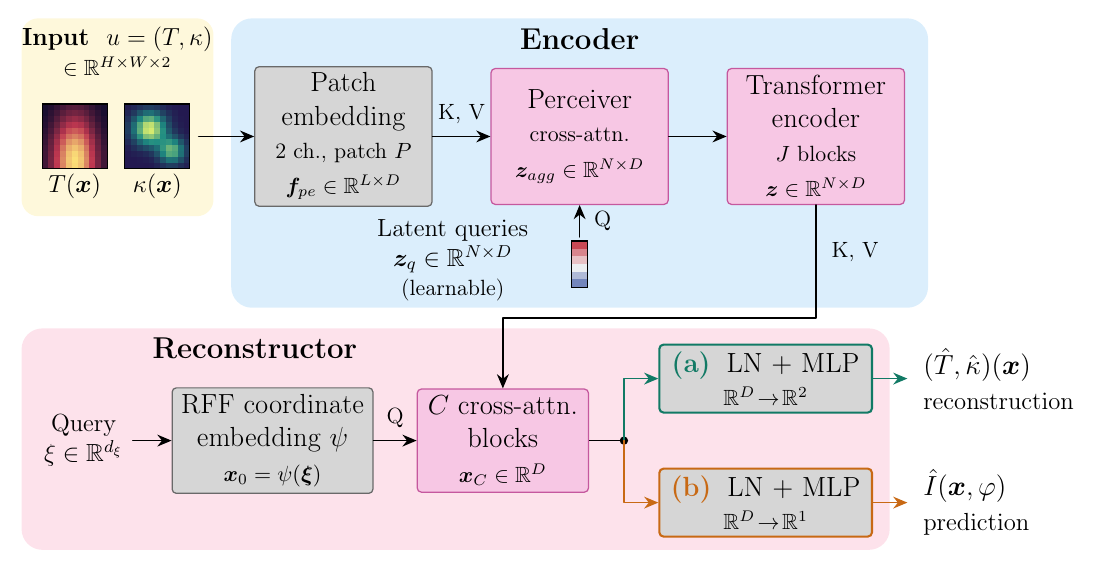}}
  \caption{\label{fig:fae}Transformer-based DeepONet. The input pair $(T,\kappa)$, given on any $H\times W$ grid, is cut into patches and embedded into $L$ tokens, $L$ following the resolution; $N$ trainable latent queries $\bm{z}_q$ collect them in one cross-attention pass, and $J$ Transformer blocks refine the result into the latent state $\bm{z}\in\mathbb{R}^{N\times D}$, whose size is the only one frozen at training time. The reconstructor embeds a query $\xi$ with random Fourier features and lets it attend to $\bm{z}$ through $C$ cross-attention blocks, so that the prediction is a nonaffine function of $\bm{z}$ and of $\xi$. The two heads correspond to the two uses of the network: (a) reconstruction of the input pair, $\xi=(x,y)$, used as a pre-training, and (b) prediction of the intensity, $\xi=(x,y,\varphi)$, which reuses the encoder weights optimised in (a).}
\end{figure}

Figure~\ref{fig:fae} illustrates the assembled architecture. In what follows, we detail the patch embedding process, the cross-attention mechanisms of both the encoder and reconstructor, the pre-training strategy, and the structural properties that distinguish this model from conventional affine baselines.

\subsubsection{From the grid to a sequence of patches.} The input pair $u = (T, \kappa)$ is given on a grid of resolution $H \times W$ where $H$ and $W$ may vary across different snapshots, and is treated as a two-channel image $\bm{f} \in \mathbb{R}^{H \times W \times 2}$. As in the classic vision transformer mechanism, it is cut into non-overlapping patches of size $P\times P$ over the domain $H\times W$; each patch is flattened into a row of $2P^{2}$ entries and projected into a vector in $\mathbb{R}^{D}$ by a single linear layer $\bm{W}_p\in\mathbb{R}^{2P^{2}\times D}$ shared by all the patches,
\begin{equation}
  \label{eq:patch}
  \bm{f}_p=\mathrm{Patch}_P(\bm{f})\,\bm{W}_p\in\mathbb{R}^{L\times D},
  \qquad L=\frac{HW}{P^{2}} .
\end{equation}
The input field is thus transformed into a sequence of $L$ tokens each of width $D$. Crucially, $L$ is the only model quantity dependent on input resolution: a finer grid lengthens the token sequence without altering the dimensionality of learnable parameters. Since the patch size $P$ is constant across resolutions, $L$ is strictly determined by the grid size $H \times W$. In this work, the patch size $P$ is chosen to divide both $H$ and $W$. Otherwise, zero padding can be applied to make both spatial dimensions divisible by $P$. Cutting the fields in this manner loses track of patch positions, which are recovered using a table of trainable positional embeddings (denoted as PE) which are learned on a reference patch grid and bilinearly interpolated to fit the current resolution,
\begin{equation}
  \label{eq:pe}
  \bm{f}_{pe}=\bm{f}_p+\mathrm{Interp}\big(\mathrm{PE},L\big),
\end{equation}
so that a single table serves every resolution~\cite{wang2025fundiff}.

\subsubsection{The Attention mechanism.} To process the sequence of embedded patches $\bm{f}_{pe}$ extracted from the input fields, both the encoder and reconstructor are built from the same operation. In a standard attention mechanism~\cite{vaswani2017attention}, given a matrix of queries $\bm{Q}\in\mathbb{R}^{n_q\times d_h}$ and matrices of keys and values $\bm{K},\bm{V}\in\mathbb{R}^{n_k\times d_h}$,
\begin{equation}
  \label{eq:attn}
  \mathrm{Attn}(\bm{Q},\bm{K},\bm{V})
  =\mathrm{softmax}\!\left(\frac{\bm{Q}\bm{K}^{\top}}{\sqrt{d_h}}\right)\bm{V} ,
\end{equation}
each of the $n_q$ output rows being a convex combination of the value rows, weighted by the similarity between the corresponding query and each key. The operation is run in parallel on $n_h$ heads of width $d_h=D/n_h$, each with its own projections, and the results are concatenated,
\begin{subequations}
\begin{align}
  \label{eq:mha}\bm{o}_i&=\mathrm{Attn}\big(\bm{q}\bm{W}^{Q}_{i},\,\bm{k}\bm{W}^{K}_{i},\,\bm{v}\bm{W}^{V}_{i}\big),\\
  \mathrm{MHA}(\bm{q},\bm{k},\bm{v})&=\big[\bm{o}_1,\dots,\bm{o}_{n_h}\big]\bm{W}^{O},
\end{align}
\end{subequations}
with $\mathrm{MSA}(\bm{q})=\mathrm{MHA}(\bm{q},\bm{q},\bm{q})$ the self-attention (MSA) obtained when the three arguments of the Multi-Head Attention (MHA) coincide. A key property of Eq.~\eqref{eq:attn} underpins our architecture. The output sequence has the same length as the query sequence, independently of the key--value sequence length. Cross-attention therefore allows information from an input sequence to be mapped to a query sequence of potentially different length. In what follows, LN denotes layer normalisation and multilayer perceptron (MLP) a two-layer feed-forward network with a nonlinear activation between the layers. All blocks use a pre-norm residual structure~\cite{xiong2020layer}.

\subsubsection{Multi-resolution encoder.\label{sec:Multi-resolution encoder}} To enable multi-resolution training, the downstream architecture must remain agnostic to the sequence length $L$, which scales with the input resolution. This is the role of the Perceiver module. Instead of letting the patch tokens attend to one another as in a standard ViT, the Perceiver introduces a set of $N$ latent tokens $\bm{z}_q\in\mathbb{R}^{N\times D}$ whose number does not follow the resolution of the input. They provide the queries of one cross-attention block whose keys and values are the patch tokens,
\begin{subequations}
\begin{align}
  \label{eq:perceiver}
  \bm{h} &= \mathrm{LN}(\bm{f}_{pe}),\\
  \bm{z}' &= \bm{z}_q+\mathrm{MHA}\big(\mathrm{LN}(\bm{z}_q),\bm{h},\bm{h}\big),\\
  \bm{z}_\mathrm{agg} &= \bm{z}'+\mathrm{MLP}\big(\mathrm{LN}(\bm{z}')\big).
\end{align}
\end{subequations}

This is the fundamental reason why the Perceiver can handle multi-resolution inputs. The aggregated state $\bm{z}_\mathrm{agg}\in\mathbb{R}^{N\times D}$ has a fixed size, whatever the length $L$ of $\bm{f}_{pe}$. The model therefore reads inputs of different resolutions, both in training and at inference.

The aggregated tokens are then refined by $J$ Transformer blocks acting on the $N$ latent tokens only,
\begin{subequations}
\begin{align}
  \label{eq:encblocks}
  \bm{z}_0 &= \mathrm{LN}(\bm{z}_\mathrm{agg}),\\
  \bm{z}'_j &= \bm{z}_{j-1}+\mathrm{MSA}\big(\mathrm{LN}(\bm{z}_{j-1})\big),\\
  \bm{z}_j &= \bm{z}'_j+\mathrm{MLP}\big(\mathrm{LN}(\bm{z}'_j)\big),
\end{align}
\end{subequations}
for $j=1,\dots,J$, the self-attention modelling the interactions between latent tokens. For simplicity, we adopt $\bm{z}=\bm{z}_J\in\mathbb{R}^{N\times D}$ for the encoder output. The depth of the network is thus dictated by a fixed sequence length $N$, which is chosen based on the desired capacity rather than the data resolution.

\subsubsection{Non-affine reconstructor.} 
The non-affine reconstructor takes the latent representation $\bm{z}$ from the encoder and evaluates the output field at an arbitrary query $\xi \in \mathbb{R}^{d_\xi}$. Following CViT~\cite{wang2025cvit}, this evaluation is made continuous via a cross-attention mechanism between the query coordinates and the encoded features. The query is first lifted into the space $\mathbb{R}^{D}$ of the latent tokens by a coordinate embedding $\psi$, for which random Fourier features and grid-based embeddings~\cite{wang2025cvit} are the two usual choices; we use the former in this work.

The embedded query $\bm{x}_0=\psi(\xi)\in\mathbb{R}^{D}$ is then passed through $C$ cross-attention blocks, in which the query comes from the coordinate and the keys and values from the encoder output,

\begin{subequations}
\begin{align}
  \bm{g} &= \mathrm{LN}(\bm{z}),\\
  \label{eq:nonlinear}
  \bm{x}'_c &= \bm{x}_{c-1}+\mathrm{MHA}\big(\mathrm{LN}(\bm{x}_{c-1}),\bm{g},\bm{g}\big),\\
  \bm{x}_c &= \bm{x}'_c+\mathrm{MLP}\big(\mathrm{LN}(\bm{x}'_c)\big),
\end{align}
\end{subequations}
for $c=1,\dots,C$; a layer normalisation and an MLP head then turn $\bm{x}_C$ into the predicted value. For the prediction of radiative intensity, the output layer applies a softplus activation function to strictly enforce the positivity of $I$, consistent with the approach of Lu \emph{et al.}~\cite{lu2026machine}. This operation introduces no additional parameters and preserves the architectural class of the model: Eq.~\eqref{eq:nonlinear} remains non-affine with respect to $\bm{z}$ regardless of its inclusion. Furthermore, since queries do not attend to one another, the reconstructor strictly operates as a pointwise evaluator, thereby maintaining the fundamental characteristic of a DeepONet.

\subsubsection{Pre-training and fine-tuning strategy}
\label{sec:2stage}
Similar to the two baselines, the network illustrated in Fig.~\ref{fig:fae} is deployed in two distinct phases, differing only in the prescribed query and output head. We frame this process as a pre-training and fine-tuning paradigm. 
In the initial pre-training phase, employing head~(a) in Fig.~\ref{fig:fae}, the query is strictly spatial, $\xi=(x,y)\in\mathbb{R}^{2}$. Here, the encoder and reconstructor are trained jointly to autoencode the input fields, returning the pair $(\hat T,\hat\kappa)(\bm{x})$. It is worth noting that extracting a latent basis, whether via PCA or a convolutional autoencoder, as in the baselines of Sec.~\ref{sec:related}, is also fundamentally a form of representation pre-training.

For the downstream prediction task, employing head~(b) in Fig.~\ref{fig:fae}, the query expands to include an angular dimension, $\xi=(x,y,\varphi)\in\mathbb{R}^{3}$, targeting the radiative intensity $\hat I(\bm{x},\varphi)$. The encoder is initialised with the weights optimised during pre-training. While these weights could be kept frozen—forcing the second stage to map from a fixed, generic representation—we elect to fine-tune the encoder alongside the new reconstructor. This strategy allows the latent representation to specialise for intensity prediction, which empirically yielded superior results.

Crucially, thanks to this neural network architecture, a single trained model can natively process inputs at varying spatial resolutions, an advantage not shared by the baseline methods of Sec.~\ref{sec:related}. Furthermore, because the reconstruction operates non-affinely upon the latent state, the proposed model fundamentally departs from the architectural class to which those standard baselines belong. The values of $P$, $J$ and $C$, the query embedding and the training settings are given in Appendix~\ref{app:train}.

\subsection{\label{subsec:floor}A computable lower bound for affine reconstructions}

This section turns Theorem 3.4 of~\citet{lanthaler2022error} into a number attached to a dataset: a bound on the training error and another on the test error. Because it holds for every encoder and approximator in Eq.~\eqref{eq:decomposition}, no architecture choice for them and no amount of training can bring a model of the class below it.

Following the notation of~\citet{lanthaler2022error}, any DeepONet can be written as
\begin{equation}
  \label{eq:decomposition}
  \widetilde{\mathcal{G}}=\mathcal{R}\circ\mathcal{A}\circ\mathcal{E},
  \qquad
  \mathcal{E}:\mathcal{X}\to\mathbb{R}^{m},\quad
  \mathcal{A}:\mathbb{R}^{m}\to\mathbb{R}^{p},\quad
  \mathcal{R}:\mathbb{R}^{p}\to\mathcal{Y},
\end{equation}
with $\mathcal{E}$ the encoder, $\mathcal{A}$ the approximator and $\mathcal{R}$ the reconstructor. In a vanilla DeepONet the reconstructor is the branch--trunk inner product
\begin{equation}
  \label{eq:reconstructor}
  \mathcal{R}(\bm\beta)=b_0+\sum_{l=1}^{p}\beta_l\gamma_l ,
\end{equation}
where $\bm\beta$ is the output of the branch net, $\bm\gamma$ that of the trunk net and $b_0$ an added bias; it is therefore \emph{affine} in $\bm\beta$. This is the only structural hypothesis the bound requires: it governs any model whose predictions lie, for every input, in a fixed affine subspace of the output space, of dimension at most $p$. Both PCA and AE baselines have this affine reconstruction structure Eq.~\eqref{eq:pcaoutput}; the reconstructor of the transformer-based DeepONet is not affine, so the bound does not govern it.

Let $\mu$ be a probability measure on $\mathcal{X}$, the space in which the input pair $(T,\kappa)$ lives. We seek to bound from below the error: 
\begin{equation}
  \label{eq:errormeasure}
  \widehat{\mathcal{E}}^{2}
  =\int_{\mathcal{X}}\int_{\mathcal{Q}}
   \big|\,\mathcal{G}(u)(\xi)-\widetilde{\mathcal{G}}(u)(\xi)\,\big|^{2}
   \,\mathrm{d}\xi\,\mathrm{d}\mu(u),
\end{equation}
with $\mathcal{G}$ the true operator and $\widetilde{\mathcal{G}}$ the model. Theorem 3.4 states that, whatever $\mathcal{E}$ and $\mathcal{A}$, if $\mathcal{R}$ is affine (as it is in Eq.~\eqref{eq:reconstructor}), then defining the covariance operator 
\begin{equation}
\label{eq:covariance}
  \Gamma_{\mathcal{G}_{\#}\mu}
  =\int_{\mathcal{Y}}\big(v-\mathbb{E}_\nu[v]\big)\otimes\big(v-\mathbb{E}_\nu[v]\big)
   \,\mathrm{d}\nu(v),
  \qquad \nu=\mathcal{G}_{\#}\mu ,
\end{equation}
and denoting by $\lambda_1\ge\lambda_2\ge\dots\ge\lambda_p\ge\dots\ge0$ the eigenvalues of that operator, we have
\begin{equation}
  \label{eq:thm34}
  \widehat{\mathcal{E}}^{2}\;\ge\;\sum_{k>p}\lambda_k .
\end{equation}

As stated, the theorem is abstract: $\mu$ is a measure on a function space and $\mathcal{G}$ the exact operator. Its mathematical content, however, is Theorem 3.9 of the same paper. Combined with the affine structure of Eq.~\eqref{eq:reconstructor}, it asks of the setting only that the output space be a separable Hilbert space and that the measure have finite second moments. Both hold here: $\mathcal{Y}=\mathbb{R}^{d}$ with the Euclidean inner product is such a space, and an empirical measure, being finitely supported, has finite second moments, so that Eq.~\eqref{eq:thm34} holds for the measures carried by a dataset.

Our errors are measured against snapshots produced by a solver, not against exact solutions: the operator at stake is the solver's, $\mathcal{G}_h:(T,\kappa)\mapsto I$, a well-defined operator since the boundary conditions are identical throughout the dataset~\cite{lu2026machine}. The dataset defined in Sec.~\ref{subsec:setup} consists of $M$ snapshots $I_m\in\mathbb{R}^{d}$ together with the associated input pairs $(T_m,\kappa_m)$, so that $I_m=\mathcal{G}_h(T_m,\kappa_m)$. To apply the bound to this dataset, we represent it by probability measures. For a point $z$, the Dirac mass $\delta_z$ is the probability measure that puts all its mass at $z$, so that $\int f\,\mathrm{d}\delta_z=f(z)$ for any function $f$. We introduce the empirical measures of the inputs and of the snapshots, which give the same weight $1/M$ to each sample,

\begin{equation}
  \widehat\mu_M=\frac{1}{M}\sum_{m=1}^{M}\delta_{(T_m,\kappa_m)},
  \qquad
  \widehat\nu_M=\frac{1}{M}\sum_{m=1}^{M}\delta_{I_m}=\mathcal{G}_{h\#}\widehat\mu_M ,
\end{equation}
the second equality holding by definition of the push-forward~\cite{peyre2025optimal}. For $\mu=\widehat\mu_M$ and $\mathcal{G}=\mathcal{G}_h$, the error Eq.~\eqref{eq:errormeasure} reads
$\frac{1}{M}\sum_{m}\|I_m-\widetilde{\mathcal{G}}(T_m,\kappa_m)\|^{2}$. This is exactly the error plotted in Fig.~\ref{fig:lower_bound} (Table~\ref{tab:accuracy} reports relative errors). The bound Eq.~\eqref{eq:thm34} holds for it with the covariance $\Gamma_{\widehat\nu_M}$ of the snapshots, defined as follows.

Denoting $\bar I=\frac{1}{M}\sum_m I_m$ and $X\in\mathbb{R}^{M\times d}$ the matrix with rows $I_m-\bar I$, the covariance operator Eq.~\eqref{eq:covariance}, taken at $\nu=\widehat\nu_M$, is the empirical covariance matrix
\begin{equation}
  \Gamma_{\widehat\nu_M}=\frac{1}{M}\sum_{m=1}^{M}
    \big(I_m-\bar I\big)\otimes\big(I_m-\bar I\big)
  =\frac{1}{M}X^{\top}X\;\in\;\mathbb{R}^{d\times d},
\end{equation}
whose nonzero eigenvalues are those of the $M\times M$ matrix $\frac{1}{M}XX^{\top}$. With $d=3.6\times10^{5}$ and $M\le4200$, the lower bound therefore costs one eigenvalue problem of the size of the dataset.
In practice, we compute the singular values of the centered data matrix $X$, whose rows are the flattened snapshots minus their mean. Squared and divided by $M$, they give the eigenvalues of the covariance matrix, sorted in decreasing order. We denote them $\lambda_{k,\mathrm{train}}$ when $X$ is built from the training set and $\lambda_{k,\mathrm{test}}$ when it is built from the test set.
The statement then bounds the training and test errors separately, each by the eigenvalues of its own dataset:
\begin{equation}
  \label{eq:floors}
  \widehat{\mathcal{E}}^{2}_{\mathrm{train}}\;\ge\;\sum_{k>p}\lambda_{k,\mathrm{train}} ,
  \qquad
  \widehat{\mathcal{E}}^{2}_{\mathrm{test}}\;\ge\;\sum_{k>p}\lambda_{k,\mathrm{test}} .
\end{equation}

Of the three models compared in Sec.~\ref{sec:results}, two are governed by Eq.~\eqref{eq:floors} and one is not. The PCA-DeepONet and the AE-DeepONet differ in the way the input is encoded, but both end in the affine reconstruction Eq.~\eqref{eq:reconstructor}, and each is therefore held above the lower bound associated with its own rank $p$; the transformer-based DeepONet, whose reconstructor is a cross-attention layer, is not. Whether that structural difference translates into a measurable one is the subject of Sec.~\ref{subsec:floorcomp}, where the three models are confronted with the lower bounds Eq.~\eqref{eq:floors}.

\section{\label{sec:results}Numerical experiments}

\subsection{\label{subsec:accuracy}Predictive performance and spectral analysis}
\subsubsection{\label{subsubsec:reconstruct}Reconstruction quality after pre-training}

We first assess the quality of the pretrained representations by comparing the reconstruction errors for $T$ and $\kappa$ obtained with PCA and the convolutional autoencoders. Whereas PCA provides a linear representation, a convolutional autoencoder can capture nonlinear structure that may require many PCA modes to represent. Following~\citet{lu2026machine}, we train two separate autoencoders for $T$ and $\kappa$, each with a latent dimension of $d=500$. Reconstruction is the only objective at this stage. The two autoencoders are trained solely to minimise the mean squared reconstruction error, using the settings given in Appendix~\ref{app:train}. The PCA basis is fitted on the same training snapshots.

\begin{widetext}
\begin{figure}[H]
  \centering
  \includegraphics[width=\textwidth]{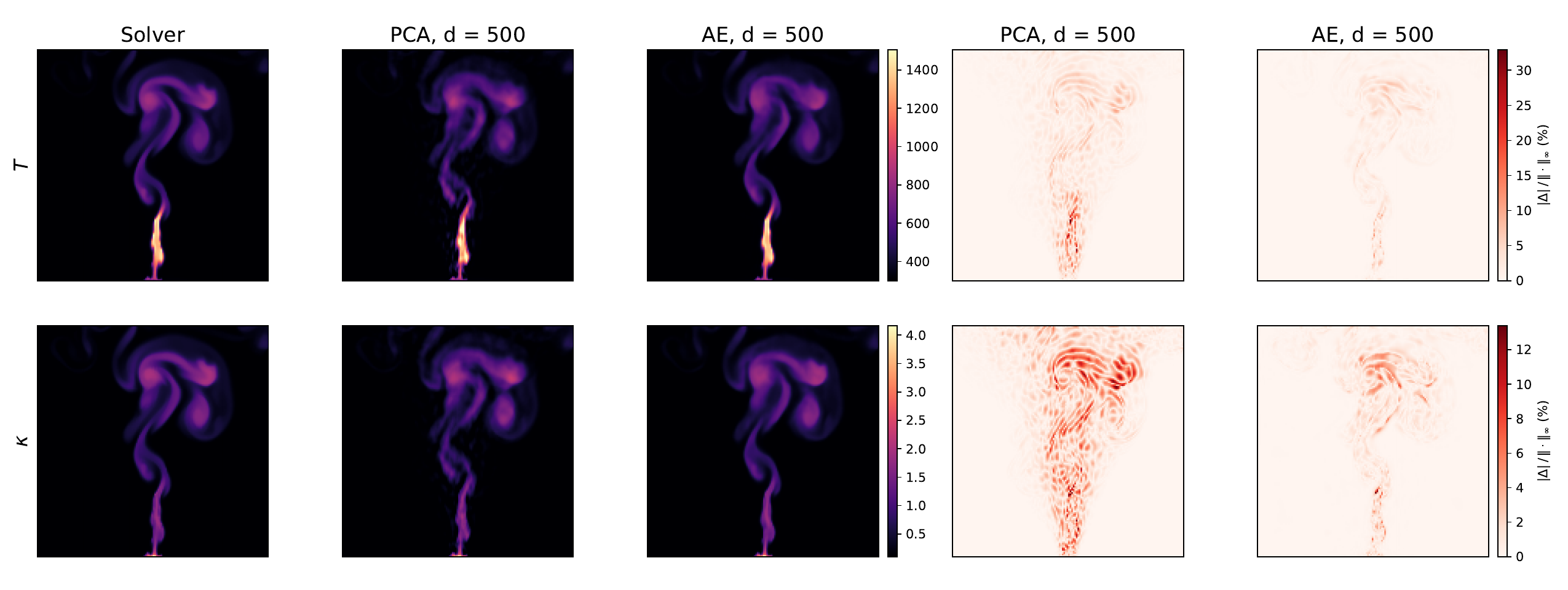}
  \caption{\label{fig:recon_compare}%
Reconstruction of one test snapshot, $T$ on top and $\kappa$ below. From left to right: the solver field, the PCA reconstruction ($d=500$), the autoencoder reconstruction ($d=500$), and the absolute error of each reconstruction at every grid point, as a percentage of the maximum of the solver field on this snapshot.}
\end{figure}
\end{widetext}

Table~\ref{tab:reconstruction} reports the relative $L^2$ reconstruction errors on the test set. At the same latent dimension of $d=500$ per field, the convolutional autoencoders reduce the error from $5.46\%$ to $1.32\%$ for $T$ and from $11.86\%$ to $3.24\%$ for $\kappa$. Figure~\ref{fig:recon_compare} illustrates this improvement on one test snapshot through the reconstructed fields and their pointwise error maps.

\begin{table}[htbp]
  \caption{\label{tab:reconstruction}%
  Relative $L^{2}$ error of the reconstruction of the input pair, on the test set. Both encoders are given the same compression dimension, $d=500$ per field.}
  \begin{ruledtabular}
  \begin{tabular}{lccc}
   & & \multicolumn{2}{c}{Reconstruction relative error (L²) (\%)} \\
  \cline{3-4}
  Encoder & $d$ ($T$ / $\kappa$) & $T$ & $\kappa$ \\
  \colrule
  PCA                       & 500 / 500 & 5.46 & 11.86 \\
  Convolutional autoencoder & 500 / 500 & 1.32 & 3.24 \\
  \end{tabular}
  \end{ruledtabular}
\end{table}

The lower reconstruction errors suggest that the autoencoders preserve the input fields more effectively than PCA at the same latent dimension. One would therefore expect AE-DeepONet to be more accurate. However, Table~\ref{tab:accuracy} shows that the improvement in input reconstruction does not translate into better prediction accuracy. These results suggest that improving the encoder alone is insufficient. Both baselines remain constrained by the same affine reconstruction structure, which restricts their predictions to an output space of dimension at most $p$, regardless of how the inputs are encoded.

\subsubsection{Accuracy of prediction}

To compare the three models, we train a reconstructor on top of each encoder, on the same training set, and report their test errors in Table~\ref{tab:accuracy}. Here, the reconstructor predicts the radiative intensity field, whereas the autoencoder decoder in Sec.~\ref{subsubsec:reconstruct} reconstructs the input field $T$ or $\kappa$. Within each model family, we vary the dimension of the representation supplied to the reconstructor. For the two baselines, this representation is a coefficient vector of dimension $p$, which limits the affine reconstruction space in Eq.~\eqref{eq:pcaoutput} to dimension at most $p$. For the transformer-based DeepONet, it consists of $N$ latent tokens of width $D$, giving $ND$ latent coordinates. The encoders of the AE- and of the transformer-based DeepONet are pretrained for the reconstruction of the input pair, then fine-tuned together with the reconstructor with the settings of Appendix~\ref{app:train}. The PCA basis, in contrast, is fitted once and then frozen.

For each model, Table~\ref{tab:accuracy} gives the relative $L^{2}$ error on the intensity $I$ and on the incident radiation $G$ of Eq. Eq.~\eqref{eq:incident_radiation}. Its last column gives the number of parameters that the prediction training actually optimises. The fixed PCA projection is excluded from the parameter count, whereas the encoder parameters are included for the other two model families.

At each rank, the two baselines are reported with two reconstructors. The affine one follows Fig.~\ref{fig:baselines}, so it is in the class of Theorem 3.4 and limited by the lower bound.~\citet{lu2026machine}, however, compose the branch--trunk inner product with a softplus to enforce the physical positivity of the intensity, which gives the reconstructor labeled ``affine + softplus'' in Table~\ref{tab:accuracy}. This adds no parameters, but the softplus is a nonlinear function of $\bm\beta$, so the resulting reconstructor is no longer affine and formally falls outside the scope of the theorem. We include this second variant, following~\citet{lu2026machine}, to assess whether softplus improves prediction accuracy in addition to enforcing positivity. According to Table~\ref{tab:accuracy}, the two variants turn out to be very close, and both are outperformed by the proposed transformer-based DeepONet. Applying softplus reduces the baseline errors by at most one third of a percentage point, with smaller reductions as $p$ increases, indicating only a modest improvement in prediction accuracy. For the remainder of the paper, we therefore consider PCA and AE-DeepONet with the affine reconstructor shown in Fig.~\ref{fig:baselines}, to which the lower bound in Theorem~3.4 applies. Results for the softplus variants are reported only in Table~\ref{tab:accuracy}. The transformer-based DeepONet retains its softplus output activation throughout the study, as its reconstructor is non-affine even without this activation.

\begin{table}[H]
  \caption{\label{tab:accuracy}
  Relative $L^{2}$ error of the prediction of the intensity $I$ and of the incident radiation $G$, on the test set. The last column gives the number of parameters optimised during the prediction training. The frozen PCA basis is therefore not counted, whereas the fine-tuned encoders of the AE-DeepONet and of the transformer-based DeepONet are counted.}
  \begin{ruledtabular}
  \begin{tabular}{lcccc}
   & &
  \multicolumn{2}{c}{Relative $L^{2}$ error (\%)} & \\
  \cline{3-4}
  Model & Reconstructor & $I$ & $G$ & Parameters \\
  \colrule
  PCA-DeepONet ($p = 128$)  & affine + softplus & 7.31  & 5.43 & 3\,075\,969 \\
  PCA-DeepONet ($p = 128$)  & affine & 7.62 & 5.73 & 3\,075\,969 \\
  PCA-DeepONet ($p = 256$)  & affine + softplus & 6.07  & 4.64 & 3\,272\,961 \\
  PCA-DeepONet ($p = 256$)  & affine & 6.20 & 4.81 & 3\,272\,961 \\
  PCA-DeepONet ($p = 512$)  & affine + softplus & 5.62  & 4.34 & 3\,666\,945 \\
  PCA-DeepONet ($p = 512$)  & affine & 5.69  & 4.49 & 3\,666\,945 \\
  \colrule
  AE-DeepONet ($p = 128$)   & affine            & 7.70 & 5.79 & 17\,607\,721 \\
  AE-DeepONet ($p = 128$)   & affine + softplus & 7.47  & 5.54 & 17\,607\,721 \\
  AE-DeepONet ($p = 256$)   & affine            & 6.62  & 5.11 & 17\,804\,713 \\
  AE-DeepONet ($p = 256$)   & affine + softplus & 6.49  & 4.92 & 17\,804\,713 \\
  AE-DeepONet ($p = 512$)   & affine            & 6.31  & 4.87 & 18\,198\,697 \\
  AE-DeepONet ($p = 512$)   & affine + softplus & 6.23  & 4.74 & 18\,198\,697 \\
  \colrule
  Transformer DeepONet ($N = 4$, $D = 32$)   & nonaffine & 14.50 & 9.72 & 96\,321 \\
  Transformer DeepONet ($N = 2$, $D = 64$)   & nonaffine & 8.11  & 5.59 & 342\,017 \\
  Transformer DeepONet ($N = 1$, $D = 128$)  & nonaffine & 5.79  & 4.05 & 1\,282\,049 \\
  Transformer DeepONet ($N = 2$, $D = 128$)  & nonaffine & 4.44  & 3.09 & 1\,282\,305 \\
  Transformer DeepONet ($N = 4$, $D = 128$)  & nonaffine & \textbf{4.40}  & \textbf{3.05} & 1\,846\,787 \\
  \end{tabular}
  \end{ruledtabular}
\end{table}

Table~\ref{tab:accuracy} allows comparisons both within and across the PCA-, AE- and transformer-based model families. Within each family, increasing the size of the latent representation improves prediction accuracy for both $I$ and $G$. The error decreases as $p$ increases for the two baselines and as $N\times D$ increases for the transformer-based DeepONet. The tested configurations suggest a greater benefit from increasing token width $D$ than token count $N$. At $N\times D=128$, widening the tokens from $D=32$ to $D=128$ lowers the error on $I$ from $14.50\%$ to $5.79\%$, while going from $N=1$ to $N=4$ at $D=128$ only brings it down to $4.40\%$.

\begin{widetext}
\begin{figure}[htbp]
  \centering
  \begin{minipage}{\textwidth}
    \textbf{(a)\; PCA-DeepONet ($p=512$), affine reconstructor}\par\smallskip
    \includegraphics[width=\linewidth]{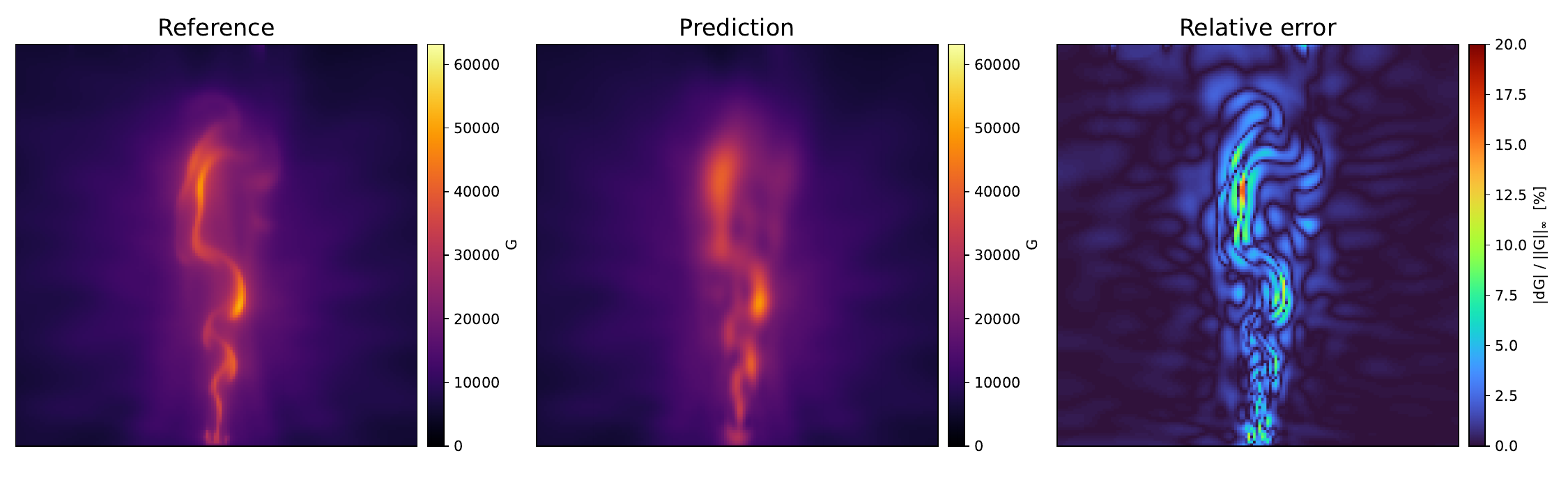}
  \end{minipage}
  \par\medskip
  \begin{minipage}{\textwidth}
    \textbf{(b)\; AE-DeepONet ($p=512$), affine reconstructor}\par\smallskip
    \includegraphics[width=\linewidth]{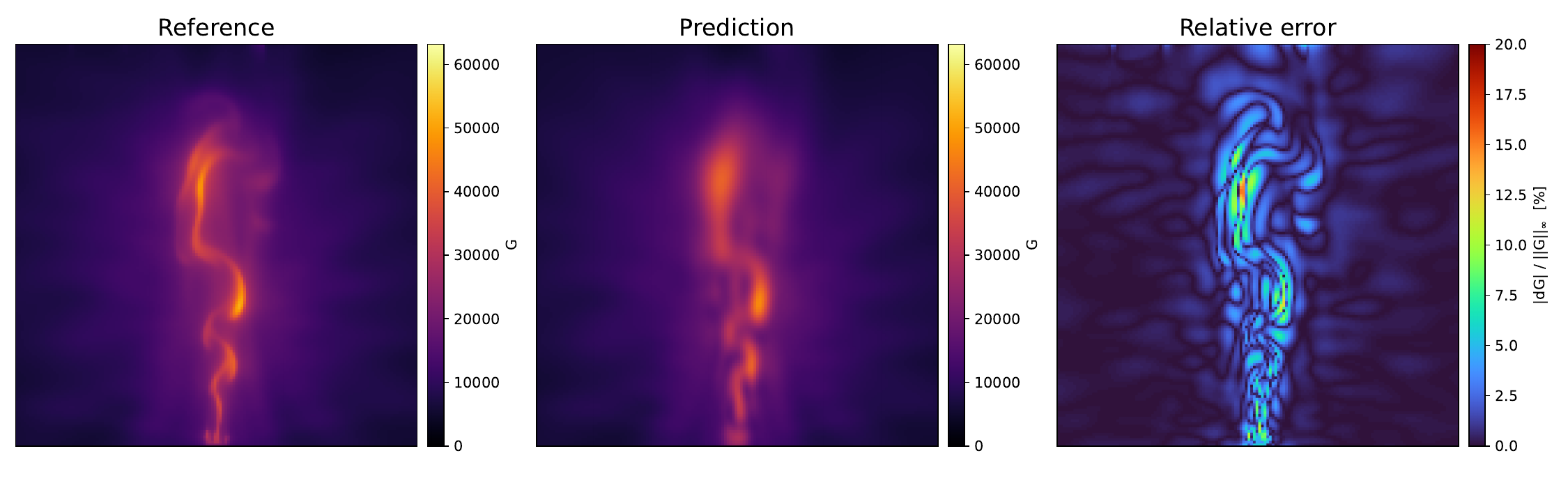}
  \end{minipage}
  \par\medskip
  \begin{minipage}{\textwidth}
    \textbf{(c)\; Transformer DeepONet ($N=4$, $D=128$), nonaffine reconstructor}\par\smallskip
    \includegraphics[width=\linewidth]{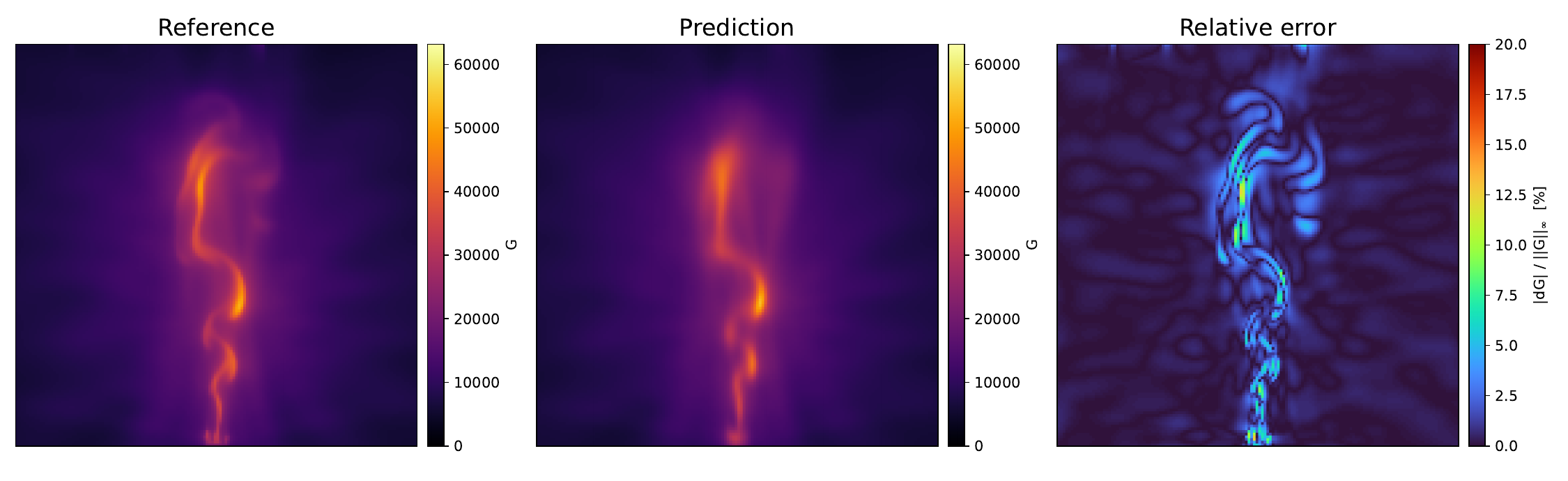}
  \end{minipage}
  \caption{\label{fig:G_compare}%
  Incident radiation $G$ on one test snapshot: solver field, prediction, and pointwise relative error. (a) PCA-DeepONet, $p=512$, affine reconstructor.  (b) AE-DeepONet, $p=512$, affine reconstructor. (c) Transformer DeepONet, $N=4$ latent tokens of width $D=128$, nonaffine reconstructor.}
\end{figure}
\end{widetext}

Across model families, the best transformer-based DeepONet configuration ($N=4$, $D=128$) achieves relative $L^2$ test errors of $4.40\%$ for $I$ and $3.05\%$ for $G$, compared with $5.62\%$ and $4.34\%$ for the best baseline. It also uses approximately half as many trainable parameters as PCA-DeepONet and one tenth as many as AE-DeepONet, both with $p=512$. The accuracy advantage is consistent across the vast majority of test snapshots. A mean error could, however, be driven by a few test snapshots on which the baselines perform poorly. We therefore compare the best affine baseline (PCA-DeepONet, $p=512$) and the best transformer-based DeepONet ($N=4$, $D=128$) snapshot by snapshot. Both have a latent dimension of $512$. The transformer-based DeepONet ($N=4$, $D=128$) has a lower error than the PCA-DeepONet ($p=512$) on $95.56\%$ of the test snapshots for $I$ and on $98.89\%$ for $G$. Its advantage is thus not confined to a few snapshots but holds across nearly the whole test set.

As an illustration, Fig.~\ref{fig:G_compare} shows the incident radiation $G$ predicted by the best model of each family on one test snapshot. All three models recover the overall plume structure, but the transformer-based DeepONet more closely reproduces the narrow, curved regions of high radiation and the local variations along the lower part of the plume. The affine baselines produce smoother, more diffuse predictions, whereas the transformer-based model exhibits smaller local discrepancies along these structures, as shown by the error maps on a common scale.

\subsubsection{\label{subsubsec:psd}Power spectral density}
Table~\ref{tab:accuracy} gives the overall accuracy of each model, but does not quantify the spatial detail each one captures, as illustrated in Fig.~\ref{fig:G_compare}. The power spectral density provides a complementary assessment. The incident radiation $G$ of Eq.~\eqref{eq:incident_radiation} is a scalar field on the mesh, so we treat it as an image, subtract its spatial mean, compute its two-dimensional discrete Fourier transform $\hat G$, and sum $|\hat G|^{2}$ over the shell of wavenumbers at distance $k$ from the origin. This is the radially averaged power spectrum used for the scale analysis of radar fields,\cite{ruzanski2011scale} with the shell average replaced by a shell sum,
\begin{equation}
  \label{eq:psd}
  P(k)=\frac{1}{n^{4}}\sum_{\bm{k}\in S_{k}}\bigl|\hat G(\bm{k})\bigr|^{2},
  \qquad S_{k}=\bigl\{\bm{k}:\ \|\bm{k}\|\in[k-\tfrac{1}{2},k+\tfrac{1}{2})\bigr\},
\end{equation}
with $n=150$ the number of mesh points per direction and $k=1,\dots,n/2$. The normalisation makes $\sum_{k}P(k)$ the spatial variance of $G$, so the curves read as a distribution of energy across scales.

\begin{widetext}
\begin{figure}[!htbp]
  \centering
  \includegraphics[width=\textwidth]{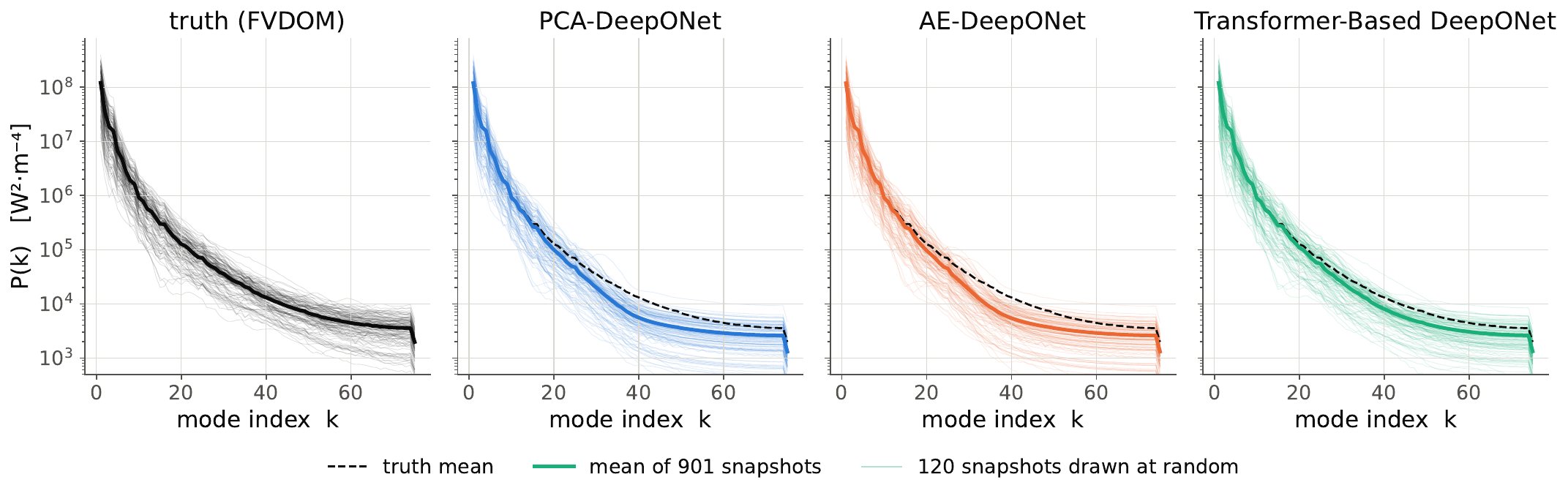}
  \caption{\label{fig:psd}%
  Power spectral density Eq.~\eqref{eq:psd} of the incident radiation $G$ on the test set. From left to right: the solver (FVDOM), the PCA-DeepONet ($p=512$), the AE-DeepONet ($p=512$) and the transformer-based DeepONet ($N=4$, $D=128$). Thin lines are $120$ test snapshots drawn at random, the thick line is the mean over the $901$ test snapshots, and the dashed line repeats the solver mean on every panel.}
\end{figure}
\end{widetext}

Figure~\ref{fig:psd} compares $P(k)$ for the solver and the best model of each family over the test set. At low wavenumbers $k$, corresponding to large-scale spatial patterns, all three models closely match the solver spectrum. Affine models with ranks of a few hundred already capture the large-scale spectral content. The curves diverge above $k\approx20$. The two baselines fall below the reference, with the gap widening as $k$ increases. This deficit in small-scale spectral power indicates excessive smoothing, consistent with the loss of fine detail visible in the error maps of Fig.~\ref{fig:G_compare}. The spectrum complements the error bound in Sec.~\ref{subsec:floor} by showing that the affine baselines lose spectral power primarily at high wavenumbers, corresponding to fine spatial detail. The transformer-based DeepONet follows the reference to higher wavenumbers. Its advantage is therefore most evident at small scales.

\subsection{\label{subsec:floorcomp}Confrontation with the lower bound}
So far, we have compared trained models, but not against theoretical error lower bounds. The transformer-based DeepONet outperforms both baselines (Table~\ref{tab:accuracy}), particularly at the small scales they smooth out (Sec.~\ref{subsubsec:psd}). These comparisons alone do not show how close the baselines are to their structural error bounds. Their errors may reflect the affine reconstruction structure, imperfect training, or both. We therefore compare them with the bound in Sec.~\ref{subsec:floor}, which gives an unavoidable structural error floor for any affine reconstructor at a fixed rank, independently of the encoder and training procedure.

\begin{figure}[htbp]
  \centering
  \includegraphics[width=\columnwidth]{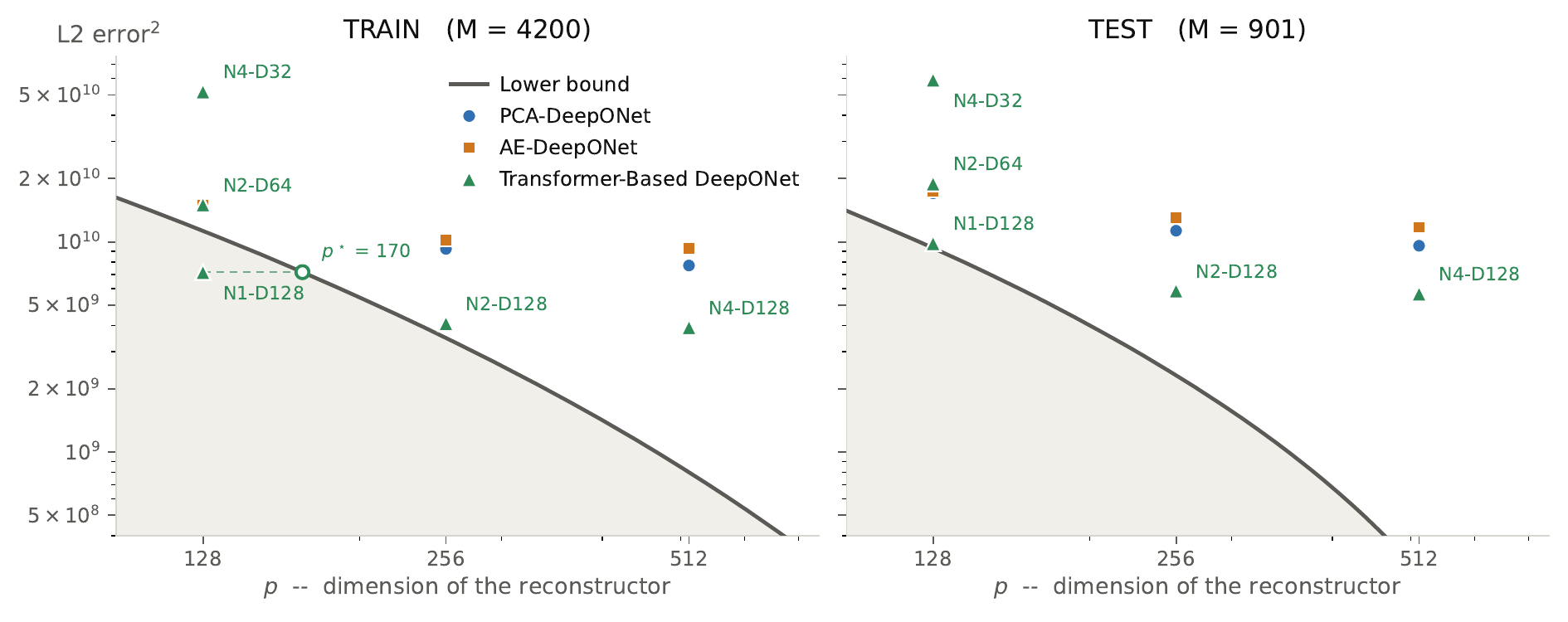}
  \caption{\label{fig:lower_bound}
  Lower bound Eq.~\eqref{eq:thm34} on the error of an affine reconstructor of rank $p$, on the training set (left, $M=4200$ snapshots) and on the test set (right, $M=901$). The shaded region is out of reach of any affine reconstructor. The markers are the models of Table~\ref{tab:accuracy}, placed at $p$ for the two baselines and at $N\times D$ for the transformer-based DeepONet. The open circle marks $p^{\star}$, the rank at which the bound comes down to the error of the best transformer-based DeepONet.}
\end{figure}

The construction in Sec.~\ref{subsec:floor} makes the theoretical lower bound in Eq.~\eqref{eq:thm34} directly computable from the dataset. For each rank $p$, it provides a lower bound on the error $\widehat{\mathcal{E}}$ of any affine reconstructor, evaluated separately on the training and test sets. This defines a curve that theoretically no PCA- or AE-DeepONet with an affine reconstructor can fall below on either the training or test dataset. Figure~\ref{fig:lower_bound} plots this bound alongside the errors of the models in Table~\ref{tab:accuracy}, each at the latent representation size used by its reconstructor ($p$ for the two baselines and $N\times D$ for the transformer-based DeepONet).

As shown in Fig.~\ref{fig:lower_bound}, the two affine models remain above their corresponding lower bounds at every evaluated rank, as expected, whereas the transformer-based DeepONet, which uses a non-affine reconstructor, achieves a training error below the rank-$128$ affine lower bound with $N=1$ and $D=128$. At this matched representation size, its empirical training error is therefore unattainable by any rank-$128$ affine reconstructor. The computed lower bound implies that either affine baseline would require a rank of at least $p^{\star}=170$ to match this error. The computed lower bound decreases more rapidly with $p$ than the model errors, leaving all tested models above the corresponding affine bound at a representation size of $512$. Its rapid decay may partly reflect finite-sample effects in the empirical covariance spectra, estimated from $4200$ training snapshots and $901$ test snapshots.

Since Theorem~3.4 holds independently of the encoder, achieving an error below the corresponding affine bound requires a non-affine reconstructor. This establishes a structural limitation of the two baselines, but does not quantify how much of the observed accuracy gap it explains. The Perceiver encoder may also contribute to the transformer-based DeepONet's performance. We therefore evaluate an ablation that retains the Perceiver encoder (see Sec.~\ref{subsec:fae}) but replaces the non-affine reconstructor with an affine one (see Sec.~\ref{subsec:baselines}), to assess the contribution of the reconstruction mechanism. Table~\ref{tab:ablation} reports it at $N=1$ and $D=128$, next to the two baselines at the same rank and to the full Transformer-based model.

\begin{table}[htbp]
  \caption{\label{tab:ablation}%
  Effect of the encoder at fixed reconstructor. The three affine models share the reconstruction Eq.~\eqref{eq:pcaoutput} of rank $p=128$ and differ only in the way the input pair is encoded. The last column gives the ratio of the measured error to the bound Eq.~\eqref{eq:floors} at the latent dimension of each model.}
  \begin{ruledtabular}
  \begin{tabular}{lccccc}
   & & & \multicolumn{2}{c}{Relative $L^{2}$ error (\%)} & \\
  \cline{4-5}
  Model & Encoder & Reconstructor & $I$ & $G$ &
  $\widehat{\mathcal{E}}/\text{bound}$ (train / test) \\
  \colrule
  PCA-DeepONet ($p=128$) & affine    & affine          & 7.62 & 5.73 & 1.15 / 1.35 \\
  AE-DeepONet ($p=128$)  & nonlinear & affine          & 7.70 & 5.79 & 1.14 / 1.36 \\
  Perceiver ($p=128$)     & Perceiver & affine          & 7.14 & 5.34 & 1.05 / 1.29 \\
  Transformer DeepONet    & Perceiver & cross-attention & 5.79 & 4.05 & 0.80 / 1.03 \\
  \end{tabular}
  \end{ruledtabular}
\end{table}

Replacing the PCA encoder with a Perceiver encoder while retaining the same affine reconstructor reduces the relative $L^{2}$ error on $I$ by only $0.48$ percentage points ($7.62\%\to7.14\%$), about a quarter of the $1.83$ percentage points separating PCA-DeepONet from the full transformer-based model. The improvement over AE-DeepONet is similarly limited to $0.56$ percentage points ($7.70\%\to7.14\%$). With the affine reconstructor retained, the Perceiver encoder reduces both the training and test error-to-bound ratios, but neither falls below one as shown in Table~\ref{tab:ablation}, consistent with Theorem~3.4. The remaining gap thus appears to come from the non-affine reconstruction through cross-attention. This ablation confirms that the main limitation of the two baselines is the affine reconstruction itself. A better encoder narrows the gap, but no encoder or amount of training can take an affine model below the bound at a fixed rank.

\subsection{\label{subsec:resolution}Generalisation across input resolutions}

The input fields supplied to a surrogate within a coupled simulation may have widely varying spatial resolutions, particularly in radiative transfer applications~\cite{bordbar2013multiscale}. The two families respond to such diversity very differently, as Fig.~\ref{fig:resolution} shows. The PCA and AE encoders considered here require inputs to be resampled onto a fixed grid, whereas the transformer-based DeepONet directly processes inputs on regular grids of varying resolution. This subsection compares the performance of the proposed DeepONet with that of the baselines on a multi-resolution dataset.

\begin{figure}[H]
  \centering
  \includegraphics[width=\columnwidth]{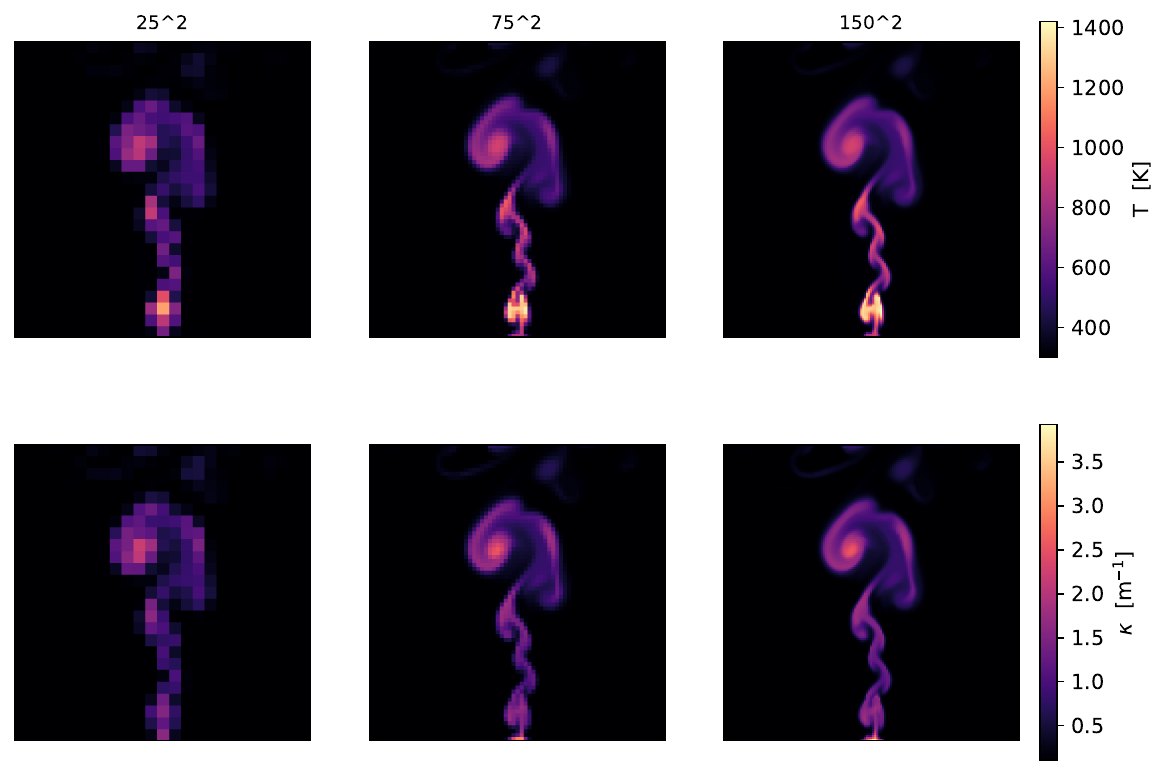}
  \caption{\label{fig:resolution_T}%
  Temperature (top) and absorption coefficient (bottom) of one snapshot at the three resolutions of the dataset, $25^{2}$, $75^{2}$ and $150^{2}$ from left to right. The same block averaging is applied to the intensity.}
\end{figure}

From the dataset of Sec.~\ref{subsec:setup}, we construct a multi-resolution dataset by block averaging, with one third of the snapshots presented at $150^{2}$, one third at $75^{2}$ and one third at $25^{2}$, as shown in Fig.~\ref{fig:resolution_T}. Block averaging is applied to $T$, $\kappa$ and $I$ alike, so a snapshot presented at $25^{2}$ contains its input pair on a $25\times25$ grid and its intensity values at the corresponding $25^{2}\times16$ query points; the $16$ angular ordinates remain unchanged. The training, validation and test split of Sec.~\ref{subsec:setup} is preserved, the resolution is randomly assigned once to each snapshot, and the three models are trained on exactly the same data.

The transformer-based DeepONet processes this dataset directly, mapping inputs at different resolutions to a fixed-size latent representation, as described in Sec.~\ref{sec:Multi-resolution encoder}. The PCA- and AE-DeepONet require each input to be interpolated onto a fixed grid before encoding. We consider two fixed input resolutions, $150^{2}$, corresponding to the solver mesh, and $50^{2}$. At $150^{2}$, coarse inputs are upsampled, whereas at $50^{2}$, inputs are either downsampled or upsampled depending on their original resolution. Each of the PCA- and AE-DeepONet baselines is therefore trained at both fixed input resolutions, yielding four variants for comparison with the single transformer-based model in Table~\ref{tab:resolution}.

At test time, the input pair $(T,\kappa)$ is provided at one of the three spatial resolutions, while predictions are always evaluated at the $150^{2}\times16$ query points of the solver mesh and compared with the corresponding solver output. The reference solution is thus the same throughout, and only the resolution of the input fields supplied to the DeepONet surrogates varies across the columns of Table~\ref{tab:resolution}. Under this common evaluation, the transformer-based DeepONet achieves the lowest relative $L^{2}$ error at all three input resolutions. The advantage is greatest at $25^{2}$, where its error is $8.81\%$ against $15.56\%$ for the best baseline. Its architecture supports multi-resolution learning by construction, mapping variable-length input sequences to a fixed-size latent representation and allowing a single model to learn directly from fields at different resolutions without the input-grid resampling required by the baselines.

\begin{table}[htbp]
  \caption{\label{tab:resolution}%
  Relative $L^{2}$ error on the intensity, for models trained on the
  multi-resolution dataset, as a function of the resolution at which the test
  input pair is supplied; the reference is the $150^{2}$ solver field in every
  case. \texttt{@} denotes the working grid of the fixed-size encoders. Both baselines use the pure affine reconstructor of Fig.~\ref{fig:baselines}.}
  \begin{ruledtabular}
  \begin{tabular}{lcccc}
   & \multicolumn{3}{c}{Input resolution} & \\
  \cline{2-4}
  $\varepsilon_{L^{2}}(I)$ (\%) & $25^{2}$ & $75^{2}$ & $150^{2}$ & All \\
  \colrule
  PCA-DeepONet @$150^{2}$  & 15.56 & 9.25 & 9.13 & 11.22\\
  PCA-DeepONet @$50^{2}$   & 15.92 & 9.98 & 9.70 & 11.78\\
  AE-DeepONet @$150^{2}$   & 15.70 & 9.45 & 9.40 & 11.42\\
  AE-DeepONet @$50^{2}$    & 15.90 & 10.13 & 9.87 & 11.88\\
  Transformer DeepONet      & \textbf{8.81} & \textbf{6.54} & \textbf{6.86} & \textbf{7.37}\\
  \end{tabular}
  \end{ruledtabular}
\end{table}

\begin{figure}[H]
  \centering
  \includegraphics[width=\columnwidth]{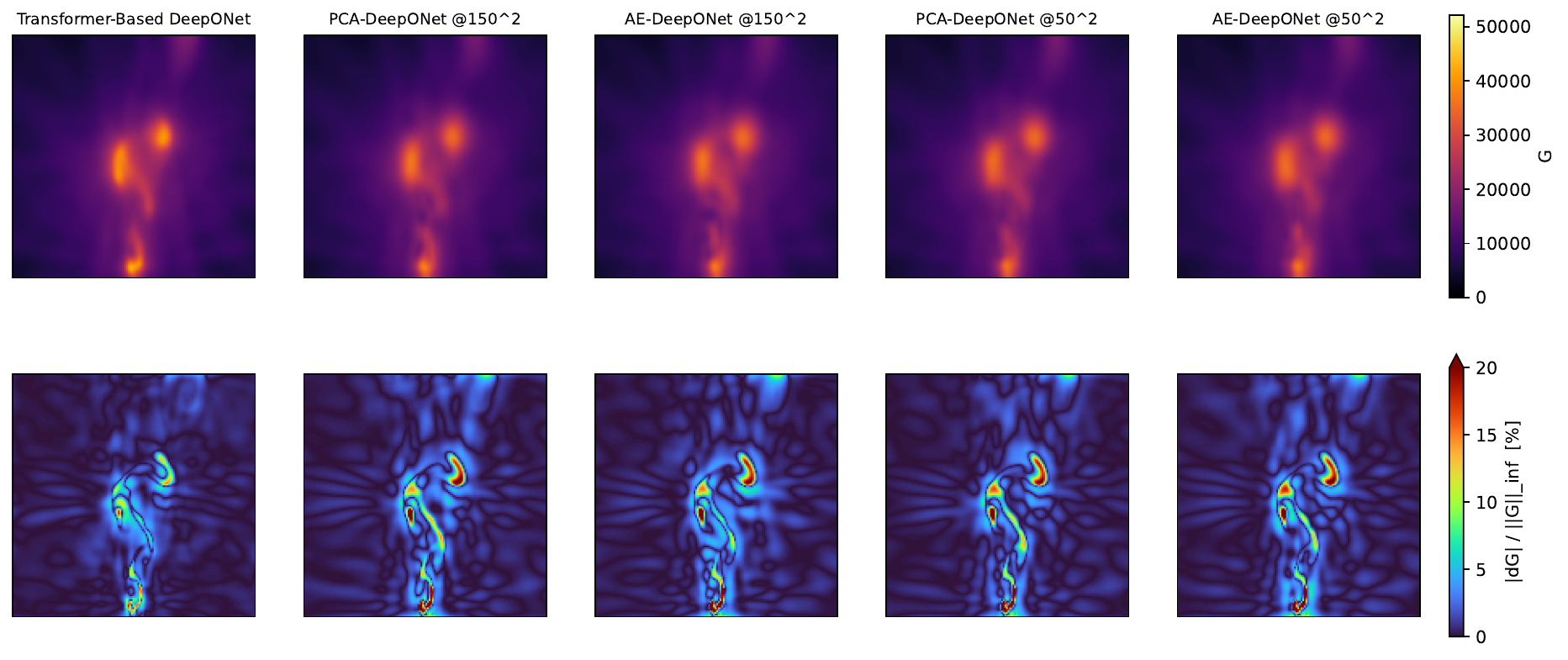}
  \caption{\label{fig:snap_res_25}%
  Test snapshot whose input pair is supplied at $25^{2}$, the prediction being requested on the $150^{2}$ solver mesh. Top row: incident radiation $G$ predicted by the model named above each panel. Bottom row: absolute error against the reference, as a percentage of $\|G\|_{\infty}$. Each row shares one colour scale.}
\end{figure}

\begin{figure}[H]
  \centering
  \includegraphics[width=\columnwidth]{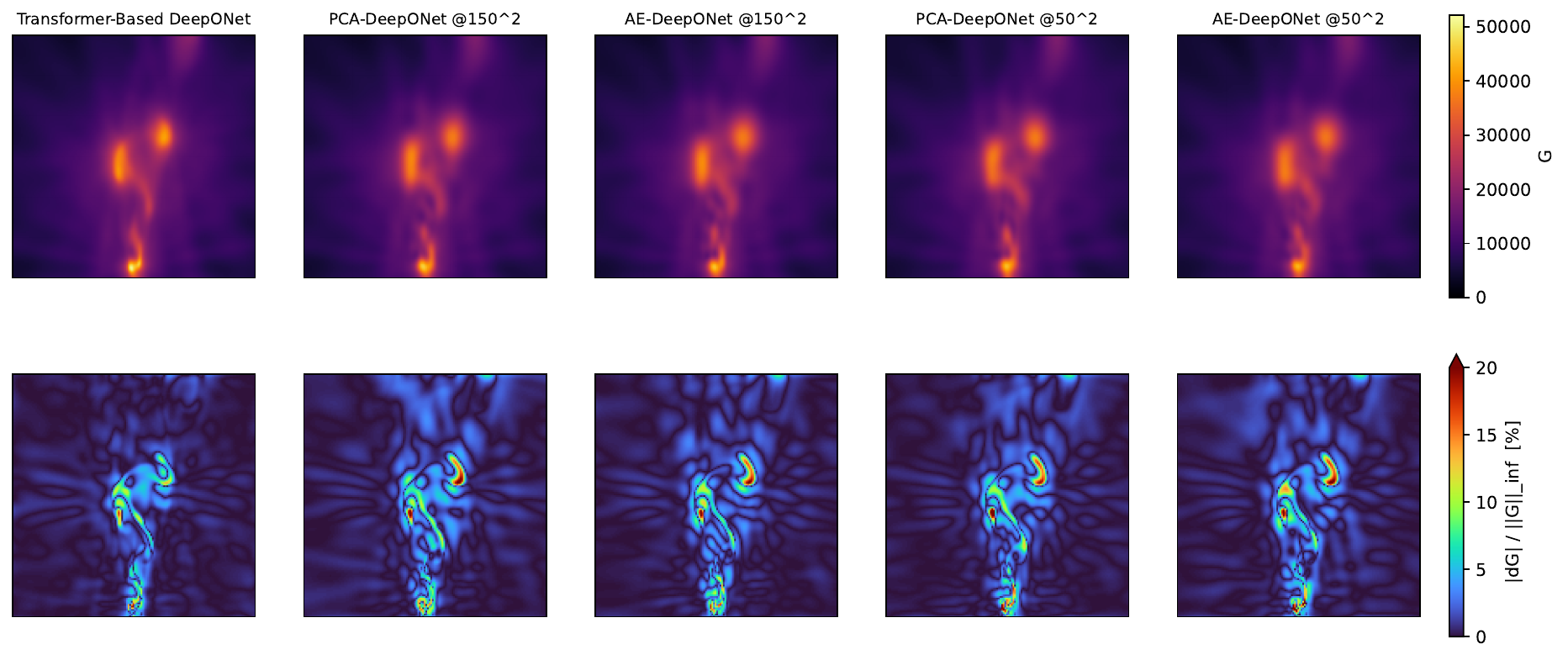}
  \caption{\label{fig:snap_res_75}%
  Same as Fig.~\ref{fig:snap_res_25}, for a test snapshot supplied at $75^{2}$.}
\end{figure}

\begin{figure}[H]
  \centering
  \includegraphics[width=\columnwidth]{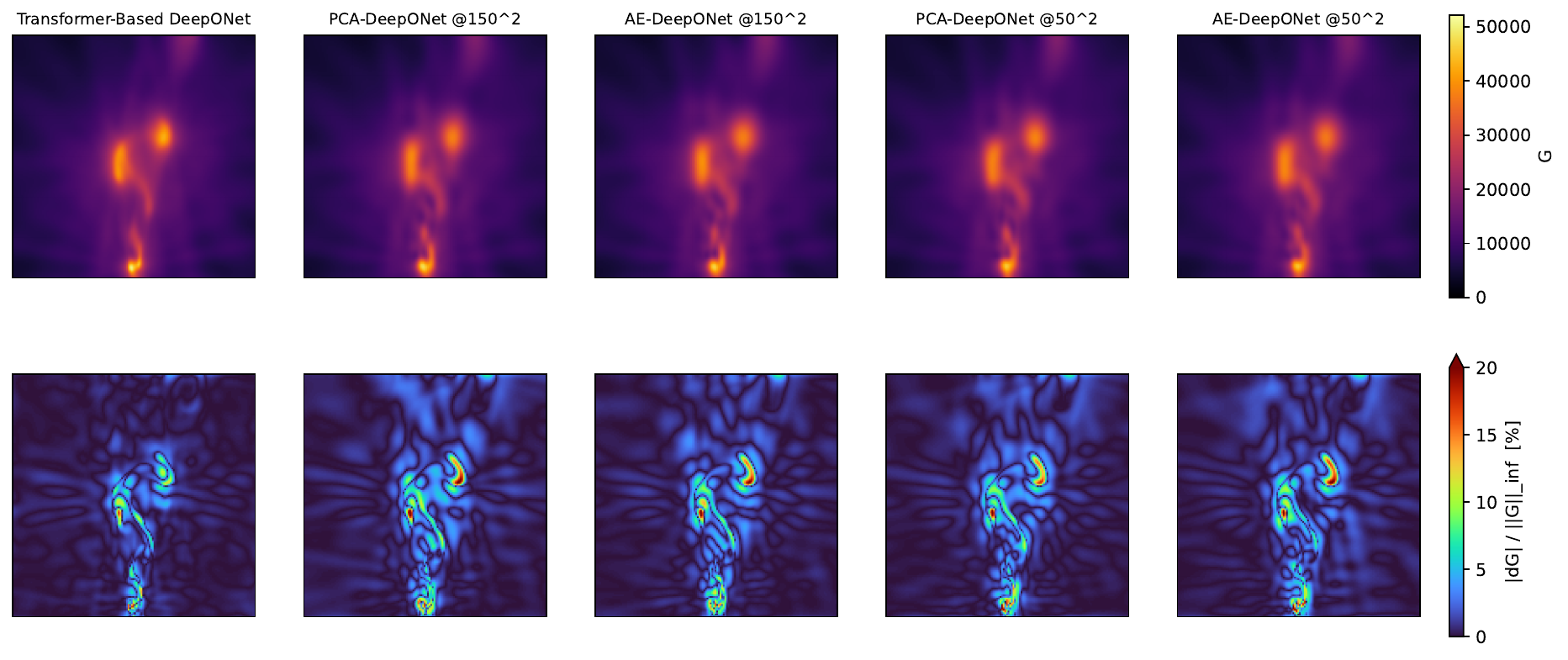}
  \caption{\label{fig:snap_res_150}%
  Same as Fig.~\ref{fig:snap_res_25}, for a test snapshot supplied at $150^{2}$,
  the resolution of the solver mesh.}
\end{figure}

Figures~\ref{fig:snap_res_25}--\ref{fig:snap_res_150} show the incident radiation $G$ of Eq.~\eqref{eq:incident_radiation} predicted by the transformer-based DeepONet and the four baseline variants for one test snapshot at each input resolution, together with the corresponding error maps. Across the three resolutions, all models reproduce the main radiation pattern, but the transformer-based DeepONet exhibits smaller and less extensive errors around the localised peaks in $G$. The baselines show more pronounced errors in these regions, with little visual difference between their two encoding grids. The advantage of the proposed model remains visible even for the $25^{2}$ inputs, consistent with the quantitative comparison in Table~\ref{tab:resolution}.

\section{\label{sec:conclusion}Conclusion}

We developed a DeepONet surrogate for the radiative transfer equation featuring two architectural departures from standard DeepONet designs. By processing input fields as patch sequences, the encoder natively accommodates varying spatial resolutions. A single trained model can therefore process multi-resolution inputs without resampling them onto a common grid. Furthermore, the reconstructor employs cross-attention, yielding a reconstruction that is non-affine in the latent representation and lies outside the class of affine reconstructors used by the baselines.
These structural advancements yield significant performance gains. At a fixed resolution, the transformer-based operator achieves superior predictive accuracy for both radiative intensity and incident radiation, requiring merely half the trainable parameters of the existing approaches. On the multi-resolution dataset, it consistently outperforms the baselines across all input resolutions, with the widest margin observed on the coarsest data. 
Our non-affine cross-attention reconstructor achieves a training error below the empirically evaluated affine bound of Lanthaler \emph{et al.} at a matched representation dimension. This identifies affine reconstruction as a structural bottleneck for the baselines at low ranks, imposing an error floor that cannot be overcome by greater encoder capacity or improved training.

For RTE modelling, the proposed surrogate combines more accurate reconstruction of fine-scale radiative structures with the ability to learn directly from temperature and absorption coefficient fields at different spatial resolutions. The proposed methodology and bound evaluation could also apply universally to any PDE dataset. In the present paper, multi-resolution training is implemented on Cartesian grids, although the Transformer architecture naturally supports training on arbitrary grid points \cite{cheng2025machine}. Temporal correlations between randomly split snapshots may also make the reported test errors optimistic, although the theoretical results remain valid and all models are evaluated using the same split. Future work could therefore focus on two key extensions: adapting the embedding mechanism to natively process unstructured and locally refined meshes, and assessing the extrapolation capability, online coupling and transient stability of this non-affine surrogate within operational CFD simulations.


\subsection*{Conflict of Interest}
The authors have no conflicts to disclose.

\subsection*{Author Contributions}
TA implemented the models, analysed the results, and prepared the draft manuscript with the contributions of all co-authors. MB, FN and JLC brought domain expertise to the project, contributing to both result analysis and manuscript development. SC developed the research idea and provided the expertise for model design. XL prepared the data. All authors reviewed and edited the manuscript.

\section*{Data Availability Statement}
The code scripts of this paper is available at \url{https://github.com/Tanguy10/Transformer-DeepONets-for-the-radiative-transfer-equation}. The data that support the findings of this study are available from the corresponding author upon reasonable request.

\section*{Acknowledgment}

 This research has been partially supported by Hi!Paris and French state funding managed by ANR under the France 2030
program, reference ANR-23-IACL-0005.
\appendix

\section{\label{app:hyper}Architectures}

This section details the three models described in Secs.~\ref{subsec:baselines} and~\ref{subsec:fae}, in the notation used there. All activations are GELU. In every model, the predicted intensity is multiplied by a fixed scale $I_s$, the median of the training intensities.

\subsection{PCA- and AE-DeepONet}

The two baselines differ only in their encoders, one per input field. In the PCA-DeepONet, each grid value of $T$ and of $\kappa$ is first normalised. The encoders $\mathcal{P}_T$ and $\mathcal{P}_\kappa$ then project the normalised field onto its first $500$ principal components.

In the AE-DeepONet, each field is read by a convolutional encoder made of two blocks. Each block applies two $3\times3$ convolutions, each followed by a group normalisation and a GELU, and ends with a max-pooling. The first block has $32$ channels and pools by a factor $2$. The second has $64$ channels and pools by a factor $5$. The $150\times150$ grid is thus reduced to $75\times75$, then to $15\times15$. A linear layer maps the resulting $64\times15^{2}=14\,400$ values to a latent vector of dimension $500$. The decoder serves only during pre-training (see Sec.~\ref{subsubsec:reconstruct}). A linear layer maps the $500$ latent coordinates back to $64\times15^{2}$ values. Two blocks then each apply a bilinear upsampling, by a factor $5$ and then $2$, followed by two $3\times3$ convolutions with $32$ channels. A final $3\times3$ convolution returns the single-channel field.

Both encoders feed the same affine reconstructor, as in Fig.~\ref{fig:baselines} and Eq.~\eqref{eq:pcaoutput}. The branch nets $B^{(T)}$ and $B^{(\kappa)}$ are MLPs with four hidden layers of width $512$ each as in \cite{lu2026machine}. Each maps the $500$ latent coordinates of its field to $\mathbb{R}^{p}$, and their outputs are combined by the Hadamard product into the coefficient vector $\bm\beta$. The trunk net $\bm\gamma$ is an MLP of the same depth and width, which maps the query $\xi=(x,y,\varphi)$ to $\mathbb{R}^{p}$. The prediction is the inner product of $\bm\beta$ and $\bm\gamma(\xi)$, plus the scalar bias $b_0$. We train this reconstructor at the ranks $p\in\{128,256,512\}$.

\subsection{Transformer-based DeepONet}

The patch embedding of Eq.~\eqref{eq:patch} uses patches of size $P=10$, so that the $150\times150$ grid yields $L=225$ tokens. The positional table of Eq.~\eqref{eq:pe} is learned on the $15\times15$ patch grid and interpolated to the other resolutions. In the Perceiver of Eq.~\eqref{eq:perceiver}, $N$ learned latent queries $\bm z_q$ gather the tokens through a single cross-attention block. Four self-attention blocks ($J=4$) then refine the latent state, as in Eq.~\eqref{eq:encblocks}. Every attention block of the model has $n_h=8$ heads of width $d_h=D/n_h$, and its MLP has a hidden width $2D$. We train five configurations, $(N,D)\in\{(4,32),(2,64),(1,128),(2,128),(4,128)\}$.

The query embedding $\psi$ first rescales $\xi$ to $[0,1]^{d_\xi}$. It then projects the rescaled query onto $64$ random Fourier frequencies, drawn once from $\mathcal{N}(0,\sigma^{2})$ with $\sigma=3$ and kept fixed. The sines and cosines of these projections give $128$ features, which a linear layer maps to $\mathbb{R}^{D}$. The reconstructor of Eq.~\eqref{eq:nonlinear} stacks $C=4$ cross-attention blocks, followed by a layer normalization and an MLP head with one hidden layer of width $D$, which maps $\mathbb{R}^{D}$ to $\mathbb{R}^{n_{\mathrm{out}}}$. The two heads of Fig.~\ref{fig:fae} differ by their query and their output. Head~(a) reads $\xi=(x,y)$ and returns $n_{\mathrm{out}}=2$ values, $\hat T$ and $\hat\kappa$. Head~(b) reads $\xi=(x,y,\varphi)$ and returns the intensity, with $n_{\mathrm{out}}=1$. Its reconstructor is newly initialised at random, while the encoder starts from its pre-trained weights.

\section{\label{app:train}Training}

Every model in this study was trained in two stages, referred to as pre-training and fine-tuning in Sec.~\ref{sec:2stage}, using the same settings.

The encoder is first pretrained on the reconstruction of the input pair $(T,\kappa)$ for $1000$ epochs, by Adam at a constant learning rate of $10^{-3}$ (the PCA is a direct fit and has no iterative training). The loss is the mean squared error of the reconstruction over the grid. For the temperature, it reads
\begin{equation}
  \label{eq:pretrainloss}
  \mathcal{L}_{T}=\frac{1}{|\mathcal{B}|}\sum_{m\in\mathcal{B}}
  \frac{1}{n}\big\|\hat T_m-T_m\big\|_{2}^{2},
\end{equation}
$\hat T_m$ being the reconstruction of $T_m$, $n=150\times150$ the number of grid points and $\mathcal{B}$ the set of snapshots of the batch. The loss $\mathcal{L}_{\kappa}$ on the absorption coefficient is defined in the same way.

The operator is then trained on the prediction of the intensity for $2000$ epochs, by Adam at a learning rate of $10^{-3}$ decayed by a factor $0.9977$ per epoch; the weights of the encoder are fine-tuned jointly at a learning rate of $10^{-4}$, under the same decay. The loss is the per-snapshot relative $L^{2}$ error, the very quantity reported in Table~\ref{tab:accuracy},
\begin{equation}
  \label{eq:trainloss}
  \mathcal{L}=\frac{1}{|\mathcal{B}|}\sum_{m\in\mathcal{B}}
  \frac{\big\|\hat I_m-I_m\big\|_{2}}{\big\|I_m\big\|_{2}}.
\end{equation}

\bibliography{refs}

\end{document}